\documentclass[letterpaper, 10 pt, conference]{packages/ieeeconf}

\IEEEoverridecommandlockouts
\usepackage[skip=2pt,font=footnotesize]{caption} 
\usepackage{multirow}
\usepackage{graphics} 
\usepackage{epsfig} 
\usepackage{times} 
\usepackage{amssymb}  
\usepackage{mathtools}
\usepackage{packages/commath}

\usepackage{amsthm}
\usepackage{graphicx}
\usepackage{subfloat}
\usepackage{subcaption}
 \usepackage[font=small]{caption}

\usepackage{array}
\newcolumntype{M}[1]{>{\centering\arraybackslash}m{#1}}

\usepackage{colortbl}
\usepackage{tabulary}

\usepackage{amsmath,amsfonts} 

\usepackage{amsmath,amssymb,amsfonts}
\usepackage{graphicx}
\DeclareGraphicsExtensions{.eps,.pdf,.png,.jpg,.JPG,.gif}
\usepackage{subfloat}
\usepackage{bm}
\usepackage{url}

\usepackage{array}
\newcolumntype{M}[1]{>{\centering\arraybackslash}m{#1}}

\usepackage{array}
\usepackage{booktabs}
\newcommand{\PreserveBackslash}[1]{\let\temp=\\#1\let\\=\temp}
\newcolumntype{C}[1]{>{\PreserveBackslash\centering}m{#1}}
\newcolumntype{R}[1]{>{\PreserveBackslash\raggedleft}p{#1}}
\newcolumntype{L}[1]{>{\PreserveBackslash\raggedright}p{#1}}

\usepackage{cite} 
\newcommand\scalemath[2]{\scalebox{#1}{\mbox{\ensuremath{\displaystyle #2}}}}

\usepackage{xcolor}

\usepackage{packages/algorithm}
\usepackage{packages/algorithmic}\usepackage{array} 
\usepackage{adjustbox}

\newcommand{\bx}{\mathbf{x}}

\newcommand{\bp}{\mathbf{p}}

\newcommand{\bz}{\mathbf{z}}

\newcommand{\bH}{\mathbf{H}}

\newcommand{\bn}{\mathbf{n}}

\newcommand{\bR}{\mathbf{R}}

\newcommand{\btheta}{\boldsymbol{\theta}}

\allowdisplaybreaks 
\renewcommand{\baselinestretch}{0.99}
\title{\huge \bf
LIC-Fusion 2.0: LiDAR-Inertial-Camera Odometry \\ with Sliding-Window Plane-Feature Tracking
}
\author{Xingxing Zuo$^{1,2}$, Yulin Yang$^3$, Patrick Geneva$^3$, Jiajun Lv$^2$, Yong Liu$^2$,  Guoquan Huang$^3$, Marc Pollefeys$^{1,4}$
    \thanks{This work was partially supported by the Department of Computer Science at ETHz and the National Natural Science Foundation of China under Grant 61836015. 
     Zuo was partially supported by the Chinese Scholarship Committee, 
    Yang was partially supported by the University of Delaware Doctoral Fellowship,
    and Geneva by the Delaware Space Grant College and Fellowship Program (NASA Grant NNX15AI19H). (Yong Liu is the corresponding author, email: yongliu@iipc.zju.edu.cn)}
	\thanks{$^1$ Department of Computer Science,  ETH Z\"{u}rich, Switzerland.}%
	\thanks{$^2$ Institute of Cyber-System and Control, Zhejiang University, Hangzhou, China.}%
	\thanks{$^3$ Robot Perception and Navigation Group, University of Delaware, Newark, DE 19716, USA.}%
	\thanks{$^4$  Microsoft Mixed Reality and Artificial Intelligence Lab, Z\"{u}rich, Switzerland.}%
}

\begin{document}
	
\maketitle

\begin{abstract}
	Multi-sensor fusion of multi-modal measurements from commodity inertial, visual and LiDAR sensors 
	to provide robust and accurate 6DOF pose estimation holds great potential in robotics and beyond. 
	In this paper, building upon our prior work (i.e., LIC-Fusion), we develop a sliding-window filter based LiDAR-Inertial-Camera odometry with online spatiotemporal calibration (i.e., LIC-Fusion 2.0), 
	which introduces a novel sliding-window plane-feature tracking for efficiently processing 3D LiDAR point clouds. 
	In particular, after motion compensation for LiDAR points by leveraging IMU data, low-curvature planar points are extracted and tracked across the sliding window.
	A novel outlier rejection criterion  is proposed in the plane-feature tracking  for high quality data association.
	Only the tracked planar points belonging to the same plane will be used for plane initialization, which makes the plane extraction  efficient and robust. 
	Moreover, we perform the observability analysis for the LiDAR-IMU subsystem and report the degenerate cases for spatiotemporal calibration using plane features. 
	While the estimation consistency and identified degenerate motions are validated in Monte-Carlo simulations, 
	different real-world experiments are also conducted to show that the proposed LIC-Fusion 2.0 outperforms its predecessor and other state-of-the-art methods.
	%
\end{abstract}

\section{Introduction and Related Work}
Accurate and robust 3D localization is essential for autonomous robots to perform high-level tasks such as autonomous driving, inspection, and delivery.
LiDAR, camera, and Inertial Measurements Unit (IMU) are among the most popular sensor choices for 3D pose estimation~\cite{Zhang2018,graeter2018limo,wan2018robust,shao2019stereo, zuo2019lic}.
Since each sensor modality has its virtues and inherent shortcomings, a proper multi-sensor fusion algorithm aiming at leveraging the ``best'' of each sensor modality is expected to have a substantial performance gain in both estimation accuracy and robustness. 
%
%
For this reason, 
Zhang and Singh~\cite{Zhang2018} proposed a graph optimization based laser-visual-inertial localization and mapping method following a multilayer processing pipeline,
in which the IMU data for prediction, a visual-inertial coupled estimator for motion estimation, and LiDAR based scan matching is integrated to further improve the motion estimation and reconstruct the map. 
In contrast to~\cite{Zhang2018}, our prior LIC-Fusion \cite{zuo2019lic} follows a lightweight filtering pipeline, which also enables spatial and temporal calibrations between the un-synchronized sensors.
In \cite{graeter2018limo}, a depth association algorithm for visual features from LiDAR measurements is developed, which is particularly suitable for autonomous driving scenarios. 
%
%
Shao et al.~\cite{shao2019stereo} fused stereo visual-inertial odometry and LiDAR scan matching within a graph optimization framework,
in which, after detecting loop closures from images, iterative closet point (ICP) of LiDAR data is performed to find the loop closure constraints.
%
%

%
\begin{figure} [!t]
	\centering
	\includegraphics[width=\columnwidth]{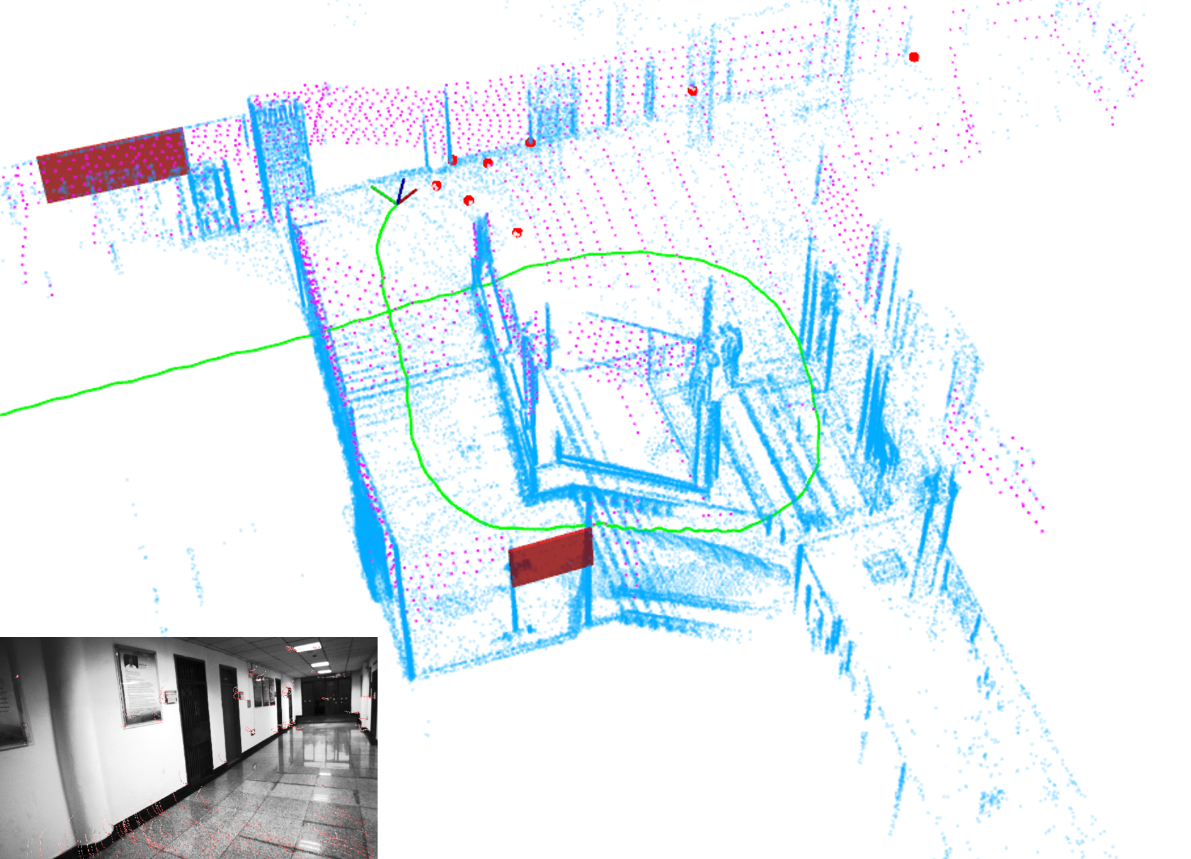}
	\caption{The proposed LIC-Fusion 2.0 with sliding-window plane-feature tracking.
		The stably tracked SLAM plane landmarks from the LiDAR and SLAM point landmarks from the camera are colored in red. 
		High curvature LiDAR points in blue, accumulated from a series of LiDAR scans, are shown to visualize the surroundings only.
		Magenta points are extracted planar points from the latest LiDAR scan.
		The estimated trajectory is marked in green. 
	}
	\label{fig:overview}
	\vspace*{-2em}
\end{figure}

Substantial research efforts have been devoted  on processing 3D LiDAR measurements to find the relative pose between two LiDAR scans.
To achieve this, ICP~\cite{besl1992method} is among the most widely used algorithms to compute the relative motion from two point clouds. 
However, traditional ICP can easily get poor results when applied on registering two 3D LiDAR scans, which have vertical sparsity and ring structure.
To cope with the sparsity in LiDAR scans, in~\cite{velas2016collar},  raw LiDAR points are converted into line segments, and the closest points from two line segments are minimized iteratively.
Similarly, in the well-known LOAM algorithm~\cite{zhang2014loam}, the registration of LiDAR scans leverage the implicit geometrical constraints (point-to-plane and point-to-line distance) to perform ``feature'' based ICP. 
This algorithm is more robust and efficient since only a few selected points with high/low curvatures are processed. 
However, both ICP and LOAM  provide constraints \textit{only} between two consecutive scans, and it is hard to accurately model the relative pose uncertainty.  
An alternative approach is to directly extract features (e.g., planes) and construct a feature-based SLAM problem~\cite{geneva2018lips}. 
However, not only is large-scale plane extraction often computationally intensive, 
but also plane-feature data association (e.g., based on Mahalanobis distance test) needs ad-hoc parameter tuning in cluttered environments.

To address these issues, building upon our prior work of LIC-fusion 1.0, 
we propose a novel plane-feature tracking algorithm to efficiently process the LiDAR measurements 
and then optimally integrated it into a sliding-window filter-based multi-sensor fusion framework
(see the overview of the system in Fig.~\ref{fig:overview}). 
In particular, after removing the motion distortion for LiDAR points, during the current sliding window, we only extract and track planar points associated with certain planes. 
Only tracked planar points will be used for plane feature initialization, which makes the plane extraction more efficient and robust. 
While abundant of works exist on observability analysis of visual-inertial systems with point features~\cite{Hesch2014TRO, Yang2019RAL}, 
we perform observability analysis for the proposed lidar-inertial-visual system and identify degenerate cases for online calibration with plane features. 
The main contributions of this work can be summarized as follows:
\begin{itemize}
	%
	\item We develop a novel sliding-window plane-feature tracking algorithm that allows for tracking 3D environmental plane features across multiple LiDAR scans within a sliding-window. This tracking algorithm is optimally integrated into our prior tightly-coupled fusion framework: LIC-Fusion~\cite{zuo2019lic}.
	For the proposed plane tracking, a novel outlier rejection criterion is advocated, which allows for robust matching by taking into account the transformation uncertainty between LiDAR frames.
	The system can model the uncertainties of LiDAR measurements reasonably, which eliminates the inconsistent-prone ICP for LiDAR scan matching.
	%
	%
	\item We perform an in-depth observability analysis of the LiDAR-inertial-camera system with plane features
	and identify the degenerate cases that cause the system to have additional unobservable directions.
	\item We conduct extensive experiments of the proposed LIC-Fusion 2.0 on a series of Monte-Carlo simulations and real-world datasets,
	which verifies both the consistency and accuracy of the proposed system.  
	%
\end{itemize}

\section{LIC-Fusion 2.0 Problem Formulation}

\subsection{State Vector}
In addition to LIC-Fusion's \cite{zuo2019lic} original state containing IMU state $\mathbf{x}_{I}$, camera clones $\mathbf{x}_{C}$, LiDAR clones $\mathbf{x}_{L}$, and spatial-temporal calibration of IMU-CAM $\mathbf{x}_{calib\_C}$ and LiDAR-IMU $\mathbf{x}_{calib\_L}$,
we store environmental visual $^{G}\bx_f$ and LiDAR landmarks ${}^{A}\bx_{\pi}$.
These features are ``long lived'' and through frequent matching can limit estimation drift. The state vector is:
\begin{align}
\label{eq:state vector}
\mathbf{x} & = 
\begin{bmatrix}
\mathbf{x}^{\top}_{I} & \mathbf{x}^{\top}_{calib\_C} & \mathbf{x}^{\top}_{calib\_L} & \mathbf{x}^{\top}_{C} & \mathbf{x}^{\top}_{L} & {}^{G}\bx_{f}^{\top} & ^{A}\bx_{\pi}^{\top}
\end{bmatrix}^{\top}
\end{align}
where
\begin{align}
\mathbf{x}_{I} & = 
\begin{bmatrix}
{}^{I_k}_G\bar{q}^{\top} & \mathbf{b}^{\top}_{g} & {}^G\mathbf{v}^{\top}_{I_k} & \mathbf{b}^{\top}_{a} & {}^G\mathbf{p}^{\top}_{I_k}
\end{bmatrix}^{\top}  \\
\label{eq:state calib C}
\mathbf{x}_{calib\_C} & = 
\begin{bmatrix}
{}^C_I\bar{q}^{\top} &  {}^C\mathbf{p}^{\top}_{I} & t_{dC}
\end{bmatrix}^{\top} \\
\label{eq:state calib L}
\mathbf{x}_{calib\_L} & = 
\begin{bmatrix}
{}^L_I\bar{q}^{\top} &  {}^L\mathbf{p}^{\top}_{I} & t_{dL}
\end{bmatrix}^{\top}\\
\mathbf{x}_{C} &=
\begin{bmatrix}
{}^{I_{c_0}}_G\bar{q}^{\top}  &   {}^G\mathbf{p}_{I_{c_0}}^{\top} &\!   \cdots  \!&  {}^{I_{c_{m-1}}}_G\bar{q}^{\top} &  {}^G\mathbf{p}_{I_{c_{m-1}}}^{\top} 
\end{bmatrix}^\top \\
\mathbf{x}_{L} &=
\begin{bmatrix}
{}^{I_{l_0}}_G\bar{q}^{\top}  &   {}^G\mathbf{p}_{I_{l_0}}^{\top} &\!   \cdots  \!&  {}^{I_{l_{n-1}}}_G\bar{q}^{\top} &  {}^G\mathbf{p}_{I_{l_{n-1}}}^{\top} 
\end{bmatrix}^\top \\
{}^G\bx_{f} &=
\begin{bmatrix}
{}^G\bp_{f_0}^{\top}  &  {}^G\bp_{f_1}^{\top}  &\!   \cdots  \!&  {}^G\bp_{f_{g-1}}^{\top} 
\end{bmatrix}^\top \\
{}^A\bx_{\pi} &=
\begin{bmatrix}
{}^A\bp_{\pi_0}^{\top}  &  {}^A\bp_{\pi_1}^{\top}  &\!   \cdots  \!&  {}^A\bp_{\pi_{h-1}}^{\top} 
\end{bmatrix}^\top
\end{align}
In the above, $\{I_k\}$ is the local IMU frame at time instant $t_k$.
${}^{I_k}_G\bar{q}$ is a unit quaternion in JPL format \cite{trawny2005indirect}, which represents 3D rotation ${}^{I_k}_G\mathbf{R}$ from $\{G\}$ to $\{I_k\}$. 
${}^G\mathbf{v}_{I_k}$, ${}^G\mathbf{p}_{I_k}$ denote the velocity and position of IMU in $\{G\}$. 
Moreover, $\mathbf{b}_{g}$ and $\mathbf{b}_{a}$ are the gyro and accelerator biases that corrupt the IMU measurements respectively.
The system error state for $x$ is defined as $ \tilde{x} = x - \hat{x}$ where $\hat{x}$ is the current estimate\footnote{ $\tilde{x}$ holds for velocity, position, bias, except for the quaternion, which follows: 
	$\bar q \simeq [\frac{1}{2}\delta \btheta^{\top} ~ 1 ]^{\top} \otimes \hat{\bar q}$,
	where $\otimes$ denotes quaternion multiplication~\cite{trawny2005indirect}, and $\delta \btheta$ is the corresponding error state.
}.
For details on the calibration parameters please see the original LIC-Fusion paper \cite{zuo2019lic}.
Additionally, we include environmental visual features, $^{G}\bp_f$, represented in the global frame of reference, and store environmental plane features represented in an anchored frame $\{A\}$.
The plane is represented by the closest point~\cite{geneva2018lips,Yang2019ICRAb}, and the anchored representation can avoid the singularity when the norm of $^{G}\bp_{\pi}$ approaches zero.
These long-lived planar features will be tracked in incoming LiDAR scans using the proposed tracking algorithm until they are lost.

\subsection{Point-to-Plane Measurement Model} \label{sec:planemeasmodel}

Considering a LiDAR planar point measurement, ${}^{L}\mathbf{p}_{f}$, that is sampled on the plane ${}^A\mathbf{p}_{\pi}$.
We can define the point-to-plane distance measurement model:
\begin{align}
\label{eq:plane meas}
\mathbf{z}_{\pi} &= 
\frac{{}^L\mathbf{p}^{\top}_{\pi} }{\norm{{}^L\mathbf{p}_{\pi}}}
({}^L\mathbf{p}_{f} - \mathbf{n}_f)
-\norm{{}^L\mathbf{p}_{\pi}}
\end{align}
where $\mathbf{n}_f\sim \mathcal{N}(\mathbf{0},\sigma_f^2\mathbf{I}_3)$.
With a slight abuse of notation, by defining ${}^Ld = \norm{{}^L\mathbf{p}_{\pi}} $ and ${}^L\mathbf{n} = {}^L\mathbf{p}_{\pi} / \norm{{}^L\mathbf{p}_{\pi}}$, a plane ${}^{A}\bp_\pi$ can be transformed into the local frame by:
\begin{align}
\begin{bmatrix}
{}^{L} \bn \\ {}^{L}d
\end{bmatrix} = \begin{bmatrix}
{}^{L}_{A}\bR & 0 \\ - {}^{A}\bp_{L}^\top & 1
\end{bmatrix} \begin{bmatrix}
{}^{A} \bn \\ {}^{A}d
\end{bmatrix}
\end{align}

\subsection{LiDAR Plane Feature Update} \label{sec:planeupdate}

Analogous to point features~\cite{li2013optimization}, we divide all the tracked plane features from the LiDAR point clouds into ``MSCKF'' and ``SLAM'' based on the track length.
Note that the sliding-window-based plane tracking will be explained in detail in Section \ref{sec:planetrack}. 
%
Given a series of LiDAR point measurements collected over the whole sliding window of the plane feature ${}^{A}\mathbf{p}_{\pi_j}$, we can linearize the measurement $\mathbf{z}^{(j)}_{f}$ in Eq. \eqref{eq:plane meas} at current estimates of ${}^{A}\mathbf{p}_{\pi_j}$ and the states $\bx$ as:
\begin{align}
\label{eq:residual ind}
\mathbf r^{(j)}_{f} & = \mathbf{0} - \mathbf{z}^{(j)}_{f} 
\simeq \mathbf H^{(j)}_{x} \tilde{\bx} + \mathbf H^{(j)}_{\pi} {}^{A}\tilde{\bp}_{\pi_j} + \bH^{(j)}_{n} \mathbf n^{(j)}_f
\end{align}
where $\mathbf{n}_f^{(j)}$ denotes the stacked noise vector. $\mathbf{H}^{(j)}_{x}$, $\mathbf H^{(j)}_{\pi}$ and $\bH^{(j)}_{n} $ are the stacked Jacobians with respect to pose states, the plane landmark and the measurement noise, respectively. 
Analytical forms of $\bH^{(j)}_{x},\bH^{(j)}_{\pi}, \bH^{(j)}_{n}$ can be found out in our companion technical report \cite{xing2020lictr}.

If ${}^{A}\mathbf{p}_{\pi_j}$ is a MSCKF plane landmark, the nullspace operation \cite{Yang2017IROS} is performed to remove the dependency on $ {}^{A}{\bp}_{\pi_j}$ by projection onto the left nullspace $\mathbf{N}$:
\begin{align} \label{eq:msckf-null-trick}
\mathbf{N}^\top\mathbf{r}^{(j)}_{f} & = 
\scalemath{0.9}{
	\mathbf{N}^\top\mathbf{H}^{(j)}_x\tilde{\mathbf{x}} + \mathbf{N}^\top\mathbf{H}^{(j)}_\pi{}^{A}\tilde{\mathbf{p}}_{\pi_j} + \mathbf{N}^\top \bH^{(j)}_{n} \mathbf{n}^{(j)}_{f}
}\\
\Rightarrow~ \mathbf{r}^{(j)}_{fo} &= \mathbf{H}^{(j)}_{xo} \tilde{\bx}+ \mathbf{n}^{(j)}_o
\label{eq:feat_update_linearized}
\end{align}
Due to the special structure that $\bH^{(j)}_{n} \bH^{(j)}_{n}{}^\top = \mathbf{I}_{n}$ the measurement covariance is still isotropic and thus the nullspace operation is valid (i.e. $\sigma_f^2 \mathbf{N}^\top \bH^{(j)}_{n} \bH^{(j)}_{n}{}^\top \mathbf{N} = \sigma_f^2\mathbf{I}_{n}$).
By stacking the residuals and Jacobians of all MSCKF plane landmarks, we obtain:
\begin{align}
\label{eq:stacked mea}
\mathbf{r}_{fo} &= \mathbf{H}_{xo} \tilde{\bx}+ \mathbf{n}_o
\end{align}
This stacked system can then update the state and covariance using the standard EKF update equations.

If ${}^{A}\mathbf{p}_{\pi_j}$ is a SLAM plane landmark that already exists in the state, we can directly update its estimate and the state using Eq. \eqref{eq:residual ind}.
To determine whether a plane feature with a long track length should be initialized into the state as a SLAM feature, we note that planes constrain the current state estimate based on their normals.
In the case that three planes that are not parallel to each other are observed, then the current state estimate can be well constrained \cite{Yang2018ICRA}.
Thus, we opt to insert ``informative'' planes whose normal directions are significantly different from the planes currently being estimated (in our implementation, we only insert planes whose normal directions have greater than ten degrees difference).
After augmenting a plane feature into the state vector, future LiDAR scans can also match to it.

\section{Sliding-Window LiDAR Plane Tracking }

\subsection{Motion Compensation for Raw LiDAR Points}\label{sec:motioncomp}

Since the raw LiDAR points are deteriorated by motion distortion, we can remove the distortion by utilizing the high-frequency IMU pose estimation.
When propagating IMU state, we save the propagated IMU poses at each timestep into a buffer, which can then be used to remove the distortion. 
Since LiDAR points occur at a higher frequency than IMU, we perform linear interpolate between each of these buffered poses to the corresponding time of each LiDAR ray.
For orientation, we perform $SO(3)$ interpolation similar to \cite{Ceriani2015IROS}, while linear interpolate between the two positions.
Using this pose, we transform all 3D points into the pose at the sweep start time, eliminating the motion distortion.

\subsection{Planar Landmark Tracking}\label{sec:planetrack}

\begin{figure} 
	\centering
	\includegraphics[width=0.7\columnwidth]{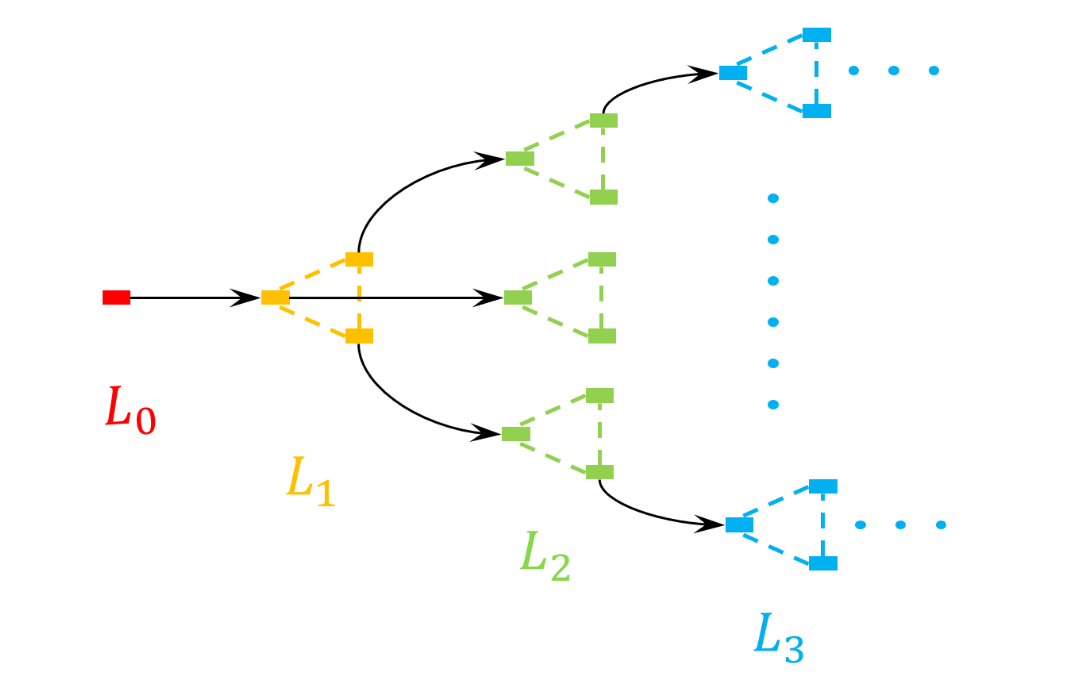}
	\caption{
		A plane landmark tracked across multiple LiDAR frames $\{L_0, L_1, L_2, L_3, \cdots \}$ within a sliding window. A planar point in the last LiDAR scan is associated with a triangle consisting of three planar points in its subsequent LiDAR scan. All the tracked planar points are assumed to be sampled from the same plane landmark.
	}
	\label{fig:tranck scans}
	\vspace*{-2.0em}
\end{figure}

We now explain how we perform temporal planar feature tracking across sequential undistorted LiDAR scans.
We first extract planar points from each LiDAR scan using the method proposed in \cite{zhang2014loam}, where low-curvature points are classified as being on some environmental planar surface.
A planar point indexed by $i$ in LiDAR frame $\{L_{a}\}$ will be tracked in the latest LiDAR frame $\{L_{b}\}$ by finding its nearest neighbour point $j$ after projection into $\{L_{b}\}$.
We then find another two points (indexed by $k, l$), which are the nearest points to $j$ on the same scan ring and the adjacent scan rings, respectively.
These three points $(j,k,l)$ are guaranteed to be non-collinear and form a planar patch corresponding to planar point $i$. 
If the distance between the projected $i$ and $j$ or distances between any two points $\in \{j,k,l\}$ are larger than a given threshold, we will reject to associate $i$ to $(j,k,l)$, and thus lose track of this planar LiDAR feature.
An overview of this approach is shown in Algorithm~\ref{alg:lidartrack} and an additional outlier rejection scheme is presented in the following section.
To prevent the reuse of information, we employ a simple strategy that a planar point can only be matched to a single common plane feature.

\begin{algorithm}
	\caption{LiDAR Plane Tracking Procedure}
	\begin{algorithmic}
		\STATE Extract planar points from $\{L_{b}\}$
		\STATE Project prior planar points from $\{L_{a}\}$ into $\{L_{b}\}$,
		find the nearest corresponding point to each in $\{L_{b}\}$.
		\FORALL{$(\mathbf{p}_i,\mathbf{p}_j) \in$ projected plane points}
		\STATE Find two closest points $\mathbf{p}_k$, $\mathbf{p}_l$ in $\{L_{b}\}$
		\STATE Ensure $\mathbf{p}_k$ scan ring is the same
		\STATE Ensure $\mathbf{p}_l$ scan rings is the adjacent
		\STATE Ensure that selected points are not already used
		\IF{$|\mathbf{p}_n-\mathbf{p}_m| < d ~~\forall (n,m) \in (i,j,k,l)$}
		\STATE Compute plane normal ${}^{b}\mathbf{n}_2$ transformed into $\{L_{a}\}$
		\STATE Compute measurement covariance matrix $\mathbf{P}_{\pi n}$
		\IF{$\chi^2(\bz_{n}, \mathbf{H}, \mathbf{P}_{\pi n}) == Pass$}
		\STATE $\mathbf{p}_j,\mathbf{p}_k,\mathbf{p}_l$ are measurements of $\mathbf{p}_i$'s plane
		\STATE $\mathbf{p}_j,\mathbf{p}_k,\mathbf{p}_l$ will be tracked into the next scan
		\ENDIF
		\ENDIF
		\ENDFOR
	\end{algorithmic}
	\label{alg:lidartrack}
\end{algorithm}

\subsection{Normal-based Plane Data Association}\label{sec:normal_assoc}

We now discuss our novel plane normal-based data association method, which rejects invalid plane associations based on the calculated plane normal.
Consider the case that we have extracted a plane on the floor next to a vertical wall.
If the tracking algorithm discussed in the previous section is used, then points that are \textit{near} the bottom of the wall would be classified as being on the same plane as floor points due to purely relying on 3D distance.
This can have huge implications on the estimation accuracy due to incorrectly saying that the wall and floor are the same plane even though their normal directions should be perpendicular to each other.

To handle this, we propose leveraging the current state uncertainty and the uncertainty of the planar points to perform a Mahalanobis distance test between the normal vectors of the candidate match.
Specifically, we have a possible planar match of the points, $({}^{L_a}\bp_{fm}, {}^{L_a}\bp_{fn}, {}^{L_a}\bp_{fo})$ in frame $\{L_a\}$, and $({}^{L_b}\bp_{fg}, {}^{L_b}\bp_{fh}, {}^{L_b}\bp_{fi})$ in frame $\{L_b\}$.
We define a synthetic measurement $\bz_{n}$ reflecting the ``parallelarity'' between the two normal vectors of each of these planes as:
\begin{align}
\label{eq:normal diff}
\bz_{n} &= \lfloor {}^{L_a} \bn_1 \rfloor {}^{L_a}_{L_b}\bR {}^{L_b} \bn_2 \\
{}^{L_a} \bn_1 &= \lfloor {}^{L_a}\bp_{fn} - {}^{L_a}\bp_{fm} \rfloor ({}^{L_a}\bp_{fo} - {}^{L_a}\bp_{fm}) \\
{}^{L_b} \bn_2 &=  \lfloor {}^{L_b}\bp_{fh} - {}^{L_b}\bp_{fg} \rfloor ({}^{L_b}\bp_{fi} - {}^{L_n}\bp_g) 
\end{align}
We can define two simplified stacked ``states'' as:
\begin{align}
\bp_{n1} &=
\begin{bmatrix} {}^{L_a}\bp_{fm}^\top & {}^{L_a}\bp_{fn}^\top & {}^{L_a}\bp_{fo}^\top \end{bmatrix}^\top \\
\bp_{n2} &=
\begin{bmatrix} {}^{L_b}\bp_{fg}^\top & {}^{L_b}\bp_{fh}^\top & {}^{L_b}\bp_{fi}^\top \end{bmatrix}^\top
\end{align}
The corresponding covariances of $\mathbf{p}_{n1}$ and $\mathbf{p}_{n2}$ can be computed from LiDAR points noises and denoted as $\mathbf{P}_{n1} = \mathbf{P}_{n2} = \sigma^2_{f} \mathbf{I}_{\mathbf{p}_{n1}}$. 
The Mahalanobis distance $d_{z}$ of $\bz_{n}$ can be computed as:
\begin{subequations}
	\begin{align}
	d_{z} & = \bz_{n}^\top \mathbf{P}_{\pi n}^{-1} \bz_{n} \\
	\mathbf{P}_{\pi n} & =
	\scalemath{0.85}{
		\frac{\partial \tilde{\mathbf{z}}_{n}}{\partial \tilde{\mathbf{p}}_{n1}} \mathbf{P}_{n1}
		\left(\frac{\partial \tilde{\mathbf{z}}_{n}}{\partial \tilde{\mathbf{p}}_{n1}}\right)^\top 
		\! +
		\frac{\partial \tilde{\mathbf{z}}_{n}}{\partial \tilde{\mathbf{p}}_{n2}} \mathbf{P}_{n2}
		\left(\frac{\partial \tilde{\mathbf{z}}_{n}}{\partial \tilde{\mathbf{p}}_{n2}}\right)^\top 
	}
	\notag \\
	&~ 
	\scalemath{0.9}{
		+ \frac{\partial \tilde{\mathbf{z}}_{n}}{\partial {}^{L_a}_{L_b}\delta\btheta } \mathbf{P}_{ori}
		\left(\frac{\partial \tilde{\mathbf{z}}_{n}}{\partial  {}^{L_a}_{L_b}\delta\btheta}\right)^\top
	}
	\end{align}
	\label{eq:normal_check_chi2_covariance}
\end{subequations}%
where $\mathbf{P}_{ori}$ is the known covariance of relative rotation ${}^{L_a}_{L_b}\bR$ based on the current EKF covariance and the Jacobians: 
\begin{subequations}
	\begin{align}
	\frac{\partial \tilde{\mathbf{z}}_{n}}{\partial \tilde{\mathbf{p}}_{n1}} &= -\lfloor {}^{L_a}_{L_b}\bR  {}^{L_b} \bn_2\rfloor \frac{\partial {}^{L_a}\tilde{\bn}_1}{\partial \tilde{\mathbf{p}}_{n1}}  \\
	\frac{\partial \tilde{\mathbf{z}}_{n}}{\partial \tilde{\mathbf{p}}_{n2}} &= \lfloor {}^{L_a} \bn_1\rfloor {}^{L_a}_{L_b}\bR  \frac{\partial {}^{L_b}\tilde{\mathbf{n}}_2}{\partial \tilde{\mathbf{p}}_{n2}}
	\end{align}
\end{subequations}
Based on the Mahalanobis distance test, we can reject incorrect temporal planar tracks. 
Note that this check can only be performed once we have more than two sequential LiDAR frames, see Fig. \ref{fig:tranck scans} for illustrating the measurements on the same plane while across multiple LiDAR frames.

\subsection{Planar Landmark Initialization}\label{sec:planeinit}

If a plane landmark ${}^{L_a}\mathbf{p}_{\pi_j}$ can be tracked across several LiDAR frames, we will initialize this plane landmark in the oldest LiDAR frame $\{L_a\}$ with all its valid planar point observations, denoted as set $\mathcal{P}_{fj}$, within the sliding window.
A planar point observation ${}^{L_x}\mathbf{p}^{(j)}_{fm_i} = {}^{L_x}\mathbf{p}^{(j)}_{f_i} + \mathbf{n}^{(j)}_{f_i}$ is the $i_{th}$ measurement in $\mathcal{P}_{fj}$, with $\mathbf{n}^{(j)}_{f_i}$ is the measurement noise.
We compute the distance between ${}^{L_x}\mathbf{p}^{(j)}_{f_i}$ and  ${}^{L_a}\mathbf{p}_{\pi_j}$ as:
\begin{align}
\label{eq:point plane dis}
\mathbf{z}^{(j)}_{f_i} & \!=\! 
\scalemath{0.8}{
	\frac{{}^{L_a}\mathbf{p}^{\top}_{\pi_j} }
	{\norm{{}^{L_a}\mathbf{p}_{\pi_j}}} 
	\left( {}^{L_a}_{L_x} \mathbf{R}
	\left({}^{L_x}\mathbf{p}^{(j)}_{fm_i} \!-\! \mathbf{n}^{(j)}_{f_i}\right) \!+\! {}^{L_a}\mathbf{p}_{L_x}
	\right)
	\!-\! \norm{{}^{L_a}\mathbf{p}_{\pi_j}}
}
\end{align} 
By stacking Eq.~\eqref{eq:point plane dis} and constructing a linear system, we can compute the initial guess for plane normal vector ${{}^{L_a}\hat{\mathbf{p}}_{\pi_j}}/{||{{}^{L_a}\hat{\mathbf{p}}_{\pi_j}}||}$ and plane distance scalar ${||{{}^{L_a}\hat{\mathbf{p}}_{\pi_j}}||}$. 
The initial guess of the plane landmark can be further refined by minimizing following cost function:
%
%
\begin{align}
{}^{L_a}{\mathbf{p}}^{*}_{\pi_j} = \mathop {\arg \min }\limits_{{}^{L_a}{\mathbf{p}}_{\pi_j}} \sum\limits_{i = 1}^n {{{\left\| \tilde{\mathbf{z}}^{(j)}_{f_i} \right\|}^2_{    \frac{\mathbf{I}}{\sigma_f^2}  }}}
\end{align}
where $n$ is the amount of observations in $\mathcal{P}_{fj}$.
%
The entire proposed LIC-Fusion 2.0 LiDAR processing pipeline can be seen in Algorithm~\ref{alg:lidarpipeline}.

\begin{algorithm}
	\caption{LIC-Fusion 2.0 LiDAR Processing Pipeline}
	\begin{algorithmic}
		\STATE \textbf{Propagation}: 
		\begin{itemize}
			\item Propagate the state forward in time by IMU measurements
			\item Buffer propagated poses for LiDAR scan motion compensation
		\end{itemize}
		\STATE \textbf{Update}: Given an incoming LiDAR Scan,
		\begin{itemize}
			\item Clone the corresponding IMU pose.
			\item Remove motion distortion for the scan as Sec.~\ref{sec:motioncomp}
			\item Extract and track planar points as Sec.~\ref{sec:planetrack}.
			\item  For SLAM plane landmarks, use the tracked planar points to compute
			the residuals $\&$ measurement Jacobians, and perform EKF update [Eq. \eqref{eq:residual ind}].
			\item For planar points that tracked across the sliding window or lost track in the current scan:
			\begin{itemize}
				\item Query its associated observations over the sliding window.
				\item Check the association validity by Mahalanobis gating test as Sec.~\ref{sec:normal_assoc}.
				\item Construct the residual vectors and the Jacobians in Eq.~\eqref{eq:point plane dis} with all the verified observations.
				\item Determine whether the plane landmark should be a SLAM landmark by checking the track length and the normal vector ``parallelity'' to the existing SLAM plane landmarks.
				\item If it should be a SLAM plane, add it to the state vector and augment the state covariance matrix. Otherwise, treat it as a MSCKF feature.
			\end{itemize}
			\item Stack the residuals and Jacobians of all MSCKF plane landmarks, and perform EKF update [Eq. \eqref{eq:stacked mea}]
		\end{itemize}
		\STATE \textbf{Management of States}: 
		\begin{itemize}
			\item SLAM plane landmarks that have lost track are marginalized out.
			\item SLAM plane landmarks anchored in the frame that needs to be marginalized are moved to the newest frame.
			\item Marginalize the cloned pose corresponding to the oldest LiDAR frame in the sliding window state.
		\end{itemize}
	\end{algorithmic}
	\label{alg:lidarpipeline}
\end{algorithm}

\section{Observability Analysis}

The observability analysis of IMU-CAM navigation system with online calibration has been studied extensively in literature~\cite{Hesch2014TRO,Yang2019TRO,Yang2019RAL}, however, the analysis for LiDAR-IMU navigation with online calibration using plane features is still missing.  
In addition, since the calibration between IMU-CAM and LiDAR-IMU calibration are relatively independent, previously identified degenerate motions for VINS calibration cannot be directly applied to LiDAR-IMU cases with plane features.  
Hence, in this paper, we focus on the subsystem of LIC-Fusion 2.0 with LiDAR-IMU only and study specifically the degenerate cases for online spatial-temporal LiDAR-IMU calibration using plane features. 
%
In particular, the observability matrix $\mathbf{M}(\mathbf{x})$ is given by:
\begin{equation}\label{eq_obs_equation}
\scalemath{.9}{
	\mathbf{M}{(\mathbf{x})} = 
	\begin{bmatrix}
	\left( \mathbf{H}_{\mathbf{x},1} \boldsymbol{\Phi}_{(1,1)}\right) ^{\top} &
	\hdots &
	\left(\mathbf{H}_{\mathbf{x},k} \boldsymbol{\Phi}_{(k,1)}\right)^{\top}
	\end{bmatrix}^{\top}
}
\end{equation}
where $\mathbf{H}_{\mathbf{x},k}$ represents the measurement Jacobians at time-step~$k$.
The right null space of $\mathbf{M}(\mathbf{x})$, denoted by $\mathbf{N}$, 
indicates the unobservable directions of the underlying system.
%

\subsection{State Vector and State Transition Matrix}

%
%
As in our previous work~\cite{Yang2019RAL}, we have already studied the observability for IMU-CAM subsystem with online calibration and point features, this analysis will only focus on LiDAR-IMU system with online calibration and plane features. 
Hence, with closest point representation for plane feature, the state vector with a plane feature and LiDAR-IMU calibration can be written as:
\begin{align}
\label{eq:plane_state}
\mathbf{x} & = \begin{bmatrix}
\mathbf{x}^{\top}_I & 
\mathbf{x}^{\top}_{calib\_L} &
{}^G\mathbf{p}^{\top}_{\pi} 
\end{bmatrix}^{\top}
\end{align}
%
The state transition matrix can be written as:
\begin{align}
\boldsymbol{\Phi}_{(k,1)} & = 
\begin{bmatrix}
\boldsymbol{\Phi}_{I} & \mathbf{0}_{15\times7}  & \mathbf{0}_{15\times3} \\
\mathbf{0}_{7\times15} & \boldsymbol{\Phi}_{calib\_L}  & \mathbf{0}_{7\times3} \\
\mathbf{0}_{3\times15}   & \mathbf{0}_{3\times7}     & \boldsymbol{\Phi}_{\pi}
\end{bmatrix}
\end{align}
Where $\boldsymbol{\Phi}_{I}$ denotes the IMU state transition matrix~\cite{Hesch2014TRO}. $\boldsymbol{\Phi}_{calib\_L} = \mathbf{I}_7$ and $\boldsymbol{\Phi}_{\pi} = \mathbf{I}_3$. 
Note that without loss of generality, we represent the plane feature in the global frame $\{G\}$.
We only consider one plane in our state vector, for the more planes cases please refer to our technical report~\cite{xing2020lictr}. 

\subsection{Measurement Jacobians and Observability Matrix}
Following the observability methodology in \cite{Hesch2014TRO}, we construct the $k$-th block of the observability matrix as:
\begin{align}
\mathbf{M}_{k} & 
= 
\mathbf{H}_{\pi}
\begin{bmatrix}
{}^L_I\mathbf{R}{^I_G\hat{\mathbf{R}}} & \mathbf{0}_{3\times1} \\
\mathbf{0}_{1\times 3} & 1
\end{bmatrix}
\times
\notag
\\
%
&
\scalemath{0.9}{
	\begin{bmatrix}
	\Gamma_{\pi 11} & \mathbf{0}_3 & \mathbf{0}_3 & \Gamma_{\pi 14} & \mathbf{0}_3 & \Gamma_{\pi 16} 
	& \mathbf{0}_3 & \Gamma_{\pi 18} & \Gamma_{\pi 19}\\
	\Gamma_{\pi 21} & {}^G\mathbf{n}^{\top} & {}^G\mathbf{n}^{\top}\Delta t_k & \Gamma_{\pi 24} & \Gamma_{\pi 25} & \Gamma_{\pi 26} 
	& \Gamma_{\pi 27} & \Gamma_{\pi 28} & \Gamma_{\pi 29}
	\end{bmatrix}
}
\notag
\end{align}
where $\Gamma_{\pi ij}, i\in\{1,2\}, j\in\{1\ldots 9\}$ can be found in~\cite{xing2020lictr}.  
For LiDAR aided INS, if the state vector contains IMU state, spatial/temporal LiDAR-IMU calibration and a plane feature, the system will have at least 7 unobservable directions as $\mathbf{N}^{(\pi)}$. 
\begin{align}
\mathbf{N}^{(\pi)} 
& = 
\begin{bmatrix}
\mathbf{N}^{(\pi)}_1 & 
\mathbf{N}^{(\pi)}_{2:4} &
\mathbf{N}^{(\pi)}_{5:6} & 
\mathbf{N}^{(\pi)}_7
\end{bmatrix}
\\
& = 
\scalemath{0.85}{
	\begin{bmatrix}
	{}^{I_1}_G\hat{\mathbf{R}}{}^G\mathbf{g}  &  \mathbf{0}_{3} & \mathbf{0}_{3\times 1} & \mathbf{0}_{3\times 1} & {}^{I_1}_G\hat{\mathbf{R}}{}^G\hat{\mathbf{n}}_{\pi} 
	\\
	-\lfloor {}^G\hat{\mathbf{p}}_{I_1} \rfloor {}^G\mathbf{g} & {}^G\hat{\mathbf{R}}_{\pi} & 
	\mathbf{0}_{3\times 1} &  \mathbf{0}_{3\times 1} &  \mathbf{0}_{3\times 1} 
	\\
	-\lfloor {}^G\hat{\mathbf{v}}_{I_1}\rfloor {}^G\mathbf{g} & \mathbf{0}_{3} & {}^G\hat{\mathbf{n}}^{\perp}_{1} & {}^G\hat{\mathbf{n}}^{\perp}_2  &  \mathbf{0}_{3\times 1}  
	\\
	\mathbf{0}_{13\times 1} & \mathbf{0}_{13\times 3} & \mathbf{0}_{13\times 1}   & \mathbf{0}_{13\times 1}  & \mathbf{0}_{13\times 1} 
	\\
	-\lfloor {}^G\hat{d}_{\pi} {}^G\hat{\mathbf{n}}_{\pi} \rfloor {}^G\mathbf{g} &  {}^G\hat{\mathbf{n}}_{\pi}\mathbf{e}^{\top}_3 & \mathbf{0}_{3\times 1} &  \mathbf{0}_{3\times 1} &  \mathbf{0}_{3\times 1}
	\end{bmatrix}  
}
\notag
\end{align}
where ${}^G\mathbf{R}_{\pi} = \begin{bmatrix} {}^G\mathbf{n}^{\perp}_{1} & {}^G\mathbf{n}^{\perp}_{2} & {}^G\mathbf{n} \end{bmatrix}$. 
%
%
%
%
%
The $\mathbf{N}^{(\pi)}_1$ relates to the global yaw around the gravity direction, $\mathbf{N}^{(\pi)}_{2:4}$ relate to the aided INS sensor platform, $\mathbf{N}^{(\pi)}_{5:6}$ relates to the velocity parallel to the plane and $\mathbf{N}^{(\pi)}_7$ relates to the rotation around the plane normal direction. 

Given 3D random motions, $\boldsymbol{\Gamma}_{\pi 16}$,  $\boldsymbol{\Gamma}_{\pi 18}$,  $\boldsymbol{\Gamma}_{\pi 26}$,  $\boldsymbol{\Gamma}_{\pi 27}$ and  $\boldsymbol{\Gamma}_{\pi 28}$ tend to have full column rank and make both the spatial and temporal calibration between LiDAR-IMU observable. 
\subsection{Degenerate Cases Analysis for LiDAR-IMU Calibration}

Given the LiDAR-IMU navigation system with plane features, the online calibration will suffer from degenerate cases that make the calibration parameters to be unobservable. 
These degenerate cases can be affected by (1) plane structure and (2) system motion. In this section, we will use one-plane case with several degenerate motions to illustrate our findings (see Table.~\ref{tab: summary of degenerate}). 
Two-plane or three-plane cases will be also included in our companion technique report. 
%
Note that the one-plane case refers to the cases when there is only one plane or all planes in the state vector are parallel. 
We have identified the following degenerate motions for the LiDAR-IMU calibration: 
\begin{itemize}
	\item If the system undergoes pure translation, the rigid transformation (including orientation and translation) between LiDAR-IMU will be unobservable with unobservable subspace as:
	\begin{align}
	\mathbf{N}^{(\pi)}_{8:11} & = 
	\scalemath{0.85}{
		\begin{bmatrix}
		\mathbf{0}_{15\times 1} & 
		\mathbf{0}_{15\times 3} \\
		{}^L_I\mathbf{R}{}^{I_1}_G\mathbf{R}{}^G\mathbf{n} & 
		\mathbf{0}_3 \\
		\mathbf{0}_{3\times1} & 
		{}^L_I\mathbf{R} {}^{I_1}_G\mathbf{R} {}^G\mathbf{R}_{\pi} \\
		0  & 0 \\
		\mathbf{0}_{3\times 1} & 
		\mathbf{e}^{\top}_3 {}^G\mathbf{n}
		\end{bmatrix}
	}
	\end{align}

	\item If rotating with the fixed axis as ${}^L\mathbf{k}$, the translation between LiDAR-IMU is not observable along the rotation axis with unobservable directions as $\mathbf{N}^{(\pi)}_{12}$. Note that if the rotation axis is perpendicular to the plane direction, we will have an extra unobservable direction $\mathbf{N}^{(\pi)}_{13}$. 
	\begin{align}
	\mathbf{N}^{(\pi)}_{12:13} & =
	\scalemath{0.85}{
		\begin{bmatrix}
		\mathbf{0}_{3\times 1}       &
		\mathbf{0}_{3\times 1}       \\
		{}^{I_1}_G\mathbf{R} {}^I_L\mathbf{R}{}^L\mathbf{k}               & 
		\mathbf{0}_{3\times 1}       \\
		\mathbf{0}_{12\times 1}      & 
		\mathbf{0}_{12\times 1}      \\
		{}^L\mathbf{k}  & 
		{}^L\mathbf{k} \\
		\mathbf{0}_{4\times 1}       &  \mathbf{0}_{4\times 1}
		\end{bmatrix}
	}
	\end{align}

	\item Similar to IMU-CAM calibration, if the system undergoes motions with constant ${}^I\boldsymbol{\omega}$ and ${}^I\mathbf{v}$ or constant ${}^I\boldsymbol{\omega}$ and $ {}^G\mathbf{a}$, the LiDAR-IMU temporal calibration will also be unobservable with unobservable directions as $\mathbf{N}^{(\pi)}_{14}$ and $\mathbf{N}^{(\pi)}_{15}$, respectively. In addition, for one-plane case, we have an extra degenerate motion (${}^G\boldsymbol{\omega} \parallel {}^G\mathbf{n}$ and ${}^G\mathbf{n}\perp {}^G\mathbf{v}_I$)
	for time offset as $\mathbf{N}^{(\pi)}_{16}$.  
	\begin{align}
	\mathbf{N}^{(\pi)}_{14:16} & \!=\!
	\scalemath{0.85}{
		\begin{bmatrix}
		\mathbf{0}_{6\times1}   & 
		\mathbf{0}_{6\times1}   &
		\mathbf{0}_{6\times1}  \\
		\mathbf{0}_{3\times1} & 
		{}^G\mathbf{a}_{I}    &
		\mathbf{0}_{3\times1} \\
		\mathbf{0}_{6\times1} &
		\mathbf{0}_{6\times1} &
		\mathbf{0}_{6\times1} \\
		{}^L_I\mathbf{R}{}^I\boldsymbol{\omega}   & 
		{}^L_I\mathbf{R}{}^I\boldsymbol{\omega}   &
		\mathbf{0}_{3\times1}                    \\
		-{}^L_I\mathbf{R}{}^I\mathbf{v} & 
		\mathbf{0}_{3\times1}           &
		\mathbf{0}_{3\times1}          \\
		-1 & -1  & 1 \\
		\mathbf{0}_{3\times1} & \mathbf{0}_{3\times1}  & \mathbf{0}_{3\times1}
		\end{bmatrix}
	}
	\end{align}
\end{itemize}
%

It can be seen that many of these degenerate motions for LiDAR-IMU coincide with the results of IMU-CAM calibration a few with additional directions.
Pure translation will cause both the orientation and translation of LiDAR-IMU extrinsic calibration unobservable, whereas for IMU-CAM calibration just the translation is unobservable.  
In addition, one-plane case will also introduce extra unobservable directions, such as $t_{dL}$, if ${}^G\boldsymbol{\omega}\parallel{}^G\mathbf{n}$ and $ {}^G\mathbf{n}\perp {}^G\mathbf{v}$. 
Note that any combination shown in Table \ref{tab: summary of degenerate} would also be degenerate. 
%
%
%
%
\begin{table}[htpb!]
	\centering
	\vspace*{-2em}
	\caption{Summary of degenerate motions for LiDAR-IMU calibration with one-plane feature.}
	\begin{tabular}{cc} \toprule
		\textbf{One Plane / Parallel Planes} & \textbf{Unobservable} \\ \midrule
		Pure Translation                                                                                            & ${}^L_I\mathbf{R}$, ${}^L\mathbf{p}_I$ \\ 
		1-axis Rotation                                                                                             & ${}^L\mathbf{p}_I $ along rotation axis  \\ 
		Constant ${}^I\omega$ and ${}^I\mathbf{v}$                                                                       & $t_{dL}$, ${}^L\mathbf{p}_I $       \\ 
		Constant ${}^I\omega$ and ${}^G\mathbf{a}$                                                                       & $t_{dL}$, ${}^L\mathbf{p}_I $       \\ 
		${}^G\boldsymbol{\omega} \parallel {}^G\mathbf{n}$ and ${}^G\mathbf{n} \perp {}^G\mathbf{v}$ &  $t_{dL} $        
		\\ 
		\bottomrule
	\end{tabular}
	\label{tab: summary of degenerate}
	\vspace*{-2.0em}
\end{table}

\section{Simulation Results}
%
%
%
\begin{table}
	\centering
	\vspace*{-2em}
	\caption{
		Simulation setup parameters. 
	}
	\label{tab:simparams}
	\begin{adjustbox}{width=\columnwidth,center}
		\begin{tabular}{ccccc} \toprule
			\textbf{Parameter} & \textbf{Value} & \textbf{Parameter} & \textbf{Value} \\ \midrule
			Cam Freq. (hz) & 10 & IMU Freq. (hz) & 200 \\
			LiDAR Freq. (hz) & 7 & LiDAR Point Noise (m) & 0.03 \\
			Gyro. White Noise & 1.6968e-04 & Gyro. Rand. Walk & 1.9393e-05 \\
			Accel. White Noise & 2.0000e-3 & Accel. Rand. Walk & 3.0000e-3 \\
			Pixel Proj. (px) & 1 & Timeoff (s) & 0.01 \\
			Rot. LtoI (rad) & 0.001 & Pos. IinL (m) & 0.01 \\
			Max Num. SLAM Point & 12 &  Max Num. SLAM Plane & 8\\
			Num. Clones Image & 11 & Num. Clones LiDAR & 8  \\\bottomrule
		\end{tabular}
	\end{adjustbox}
	\vspace*{-1em}
\end{table}
\begin{figure*}
	\centering
	\begin{subfigure}{.24\textwidth}
		\includegraphics[trim=0 0 0 0,clip,width=\linewidth]{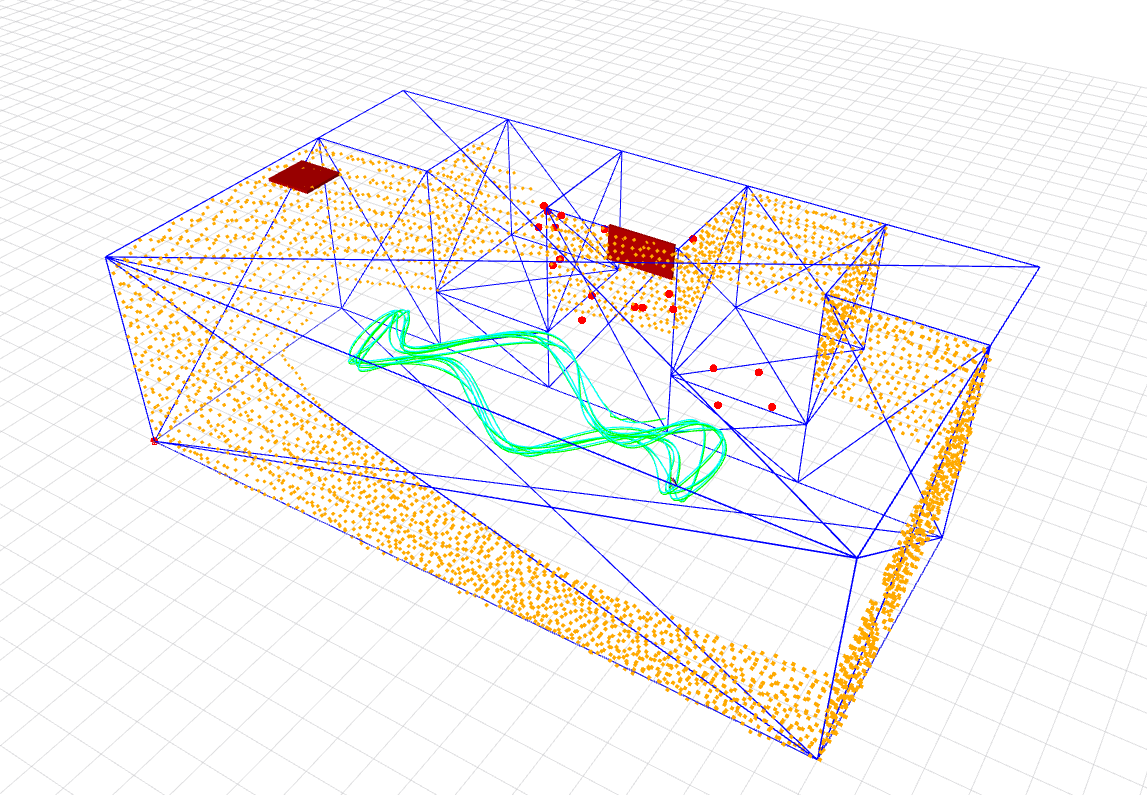}
	\end{subfigure}
	\begin{subfigure}{.24\textwidth}
		\includegraphics[trim=0 0 0 0,clip,width=\linewidth]{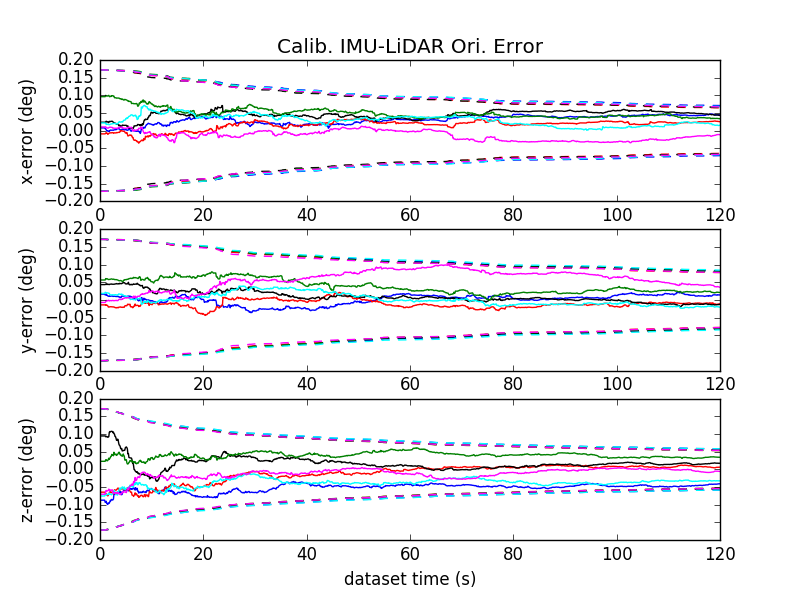}
	\end{subfigure}
	\begin{subfigure}{.24\textwidth}
		\includegraphics[trim=0 0 0 0,clip,width=\linewidth]{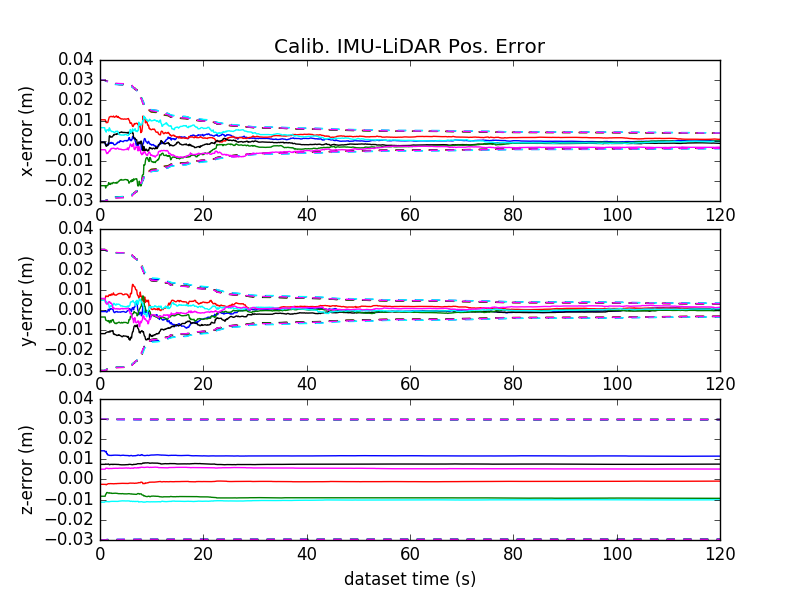}
	\end{subfigure}
	\begin{subfigure}{.24\textwidth}
		\includegraphics[trim=0 0 0 0,clip,width=\linewidth]{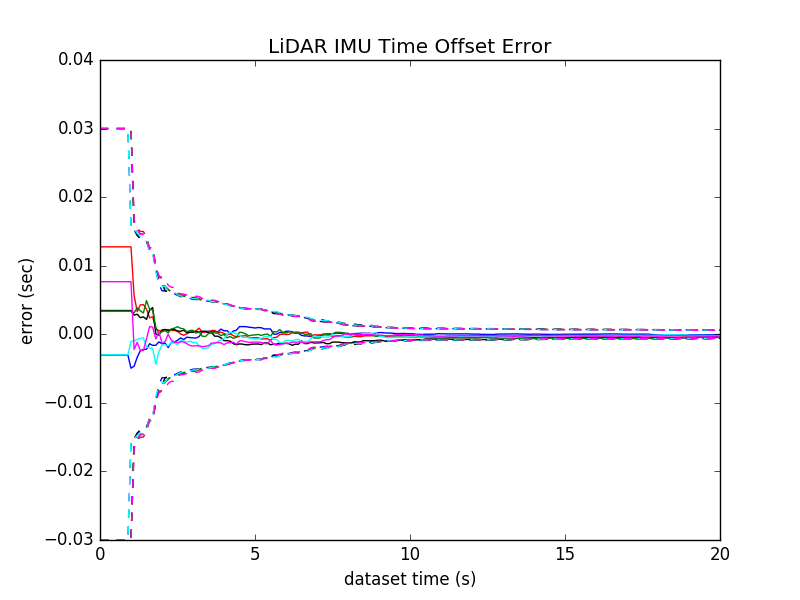}
	\end{subfigure}
	\caption{
		Left: the simulated room with structure planes (blue), 16-beam LiDAR points (yellow), SLAM point landmarks (red dot), SLAM plane landmarks (red patch), estimated (green) and ground truth (cyan) trajectories. 
		Middle and right: calibration errors and 3 sigma bounds for 6 typical runs with different initial state perturbations for 1-axis rotation motion (yaw only).
		z-component of ${}^L\mathbf{p}_I$ is not observable and does not converge at all.
	}
	\label{fig:sim_exp}
	\vspace*{-1.0em}
\end{figure*}
\begin{table}
	\centering
	\caption{
		Averaged ATE and NEES over 12 simulation runs with and without online calibration.
		Note that ``true'' means ground truth calibration while ``bad'' means the perturbed calibration and IC refers to IMU-camera subsystem only. 
	}
	\label{tab:sim_calibcompare}
	\begin{adjustbox}{width=\columnwidth,center}
		\begin{tabular}{ccccc} \toprule
			\textbf{IMU Model} & \textbf{ATE (deg)} & \textbf{ATE (m)}
			& \textbf{Ori. NEES} & \textbf{Pos. NEES} \\ \midrule
			true w/ calib & 0.118 & 0.020 & 2.210 & 0.185 \\
			%
			bad w/ calib & 0.129 & 0.021 & 2.216 & 0.221 \\
			%
			bad w/o calib &  0.148 & 0.024 & 2.677 & 0.246 \\
			true w/o calib & 0.122 & 0.021 & 2.233 & 0.208 \\
			IC true w/o calib & 0.159 & 0.027 & 2.237 & 0.314 \\
			\bottomrule
		\end{tabular}
	\end{adjustbox}
	\vspace*{-1.2em}
\end{table}
We first verify our proposed system and observability analysis in simulation. A virtual room with structural planes (Fig.~\ref{fig:sim_exp}) is simulated~\cite{Geneva2018IROS,Geneva2020ICRA}. IMU measurements, LiDAR points, sparse image features, perturbation to the initial states, and noises to measurements are generated according to configuration shown in Table~\ref{tab:simparams}. 
We first evaluate the proposed system with and without online LiDAR-IMU calibration by 12 Monte-Carlo runs, where absolute trajectory error (ATE) and Normalized Estimation Error Squared (NEES) are used to quantify accuracy and consistency, respectively. 

The results are shown in Table~\ref{tab:sim_calibcompare}, where the ``true'' denotes the system starts with ground truth calibration parameters, while the ``bad''
indicates the system is initialized with perturbed calibration. 
%
The results suggest that the proposed system with online calibration can achieve consistent and accurate pose estimation. 
In comparison, the system will output inconsistent pose estimation (much larger ATE and NEES) if it starts from perturbed initial states and runs without online calibration. 
%
Furthermore, LIC-Fusion 2.0 also outperforms its IMU-CAM (IC) subsystem (which only fuses IMU and camera measurements). 
During the simulation, 14.61 MSCKF plane landmarks and 1.60 SLAM plane landmarks are used for update on average every scan.  

We further examine a degenerate motion (1-axis rotation motion) identified for online LiDAR-IMU calibration. With the same trajectory shown in Fig.~\ref{fig:sim_exp}, we remove orientation roll and pitch changes allowing only the yaw to change.
The spatial-temporal calibration between LiDAR-IMU over 6 runs with online calibration are shown in Fig.~\ref{fig:sim_exp}. All calibration parameters except the z component of ${}^L\mathbf{p}_I$ converge nicely with shrunken uncertainty bounds. Because the sensor is rotating around z-axis (yaw only orientation), hence, the z-component of ${}^L\mathbf{p}_I$ is observable. Therefore, the results support our degenerate motion analysis, see Table~\ref{tab: summary of degenerate}.

\section{Real-world Experimental Results}
\begin{figure*} 
	\centering
	\begin{subfigure}{.19\textwidth}
		\vspace*{-0.5em}
		\includegraphics[trim=0 0 0 0,clip,width=\linewidth]{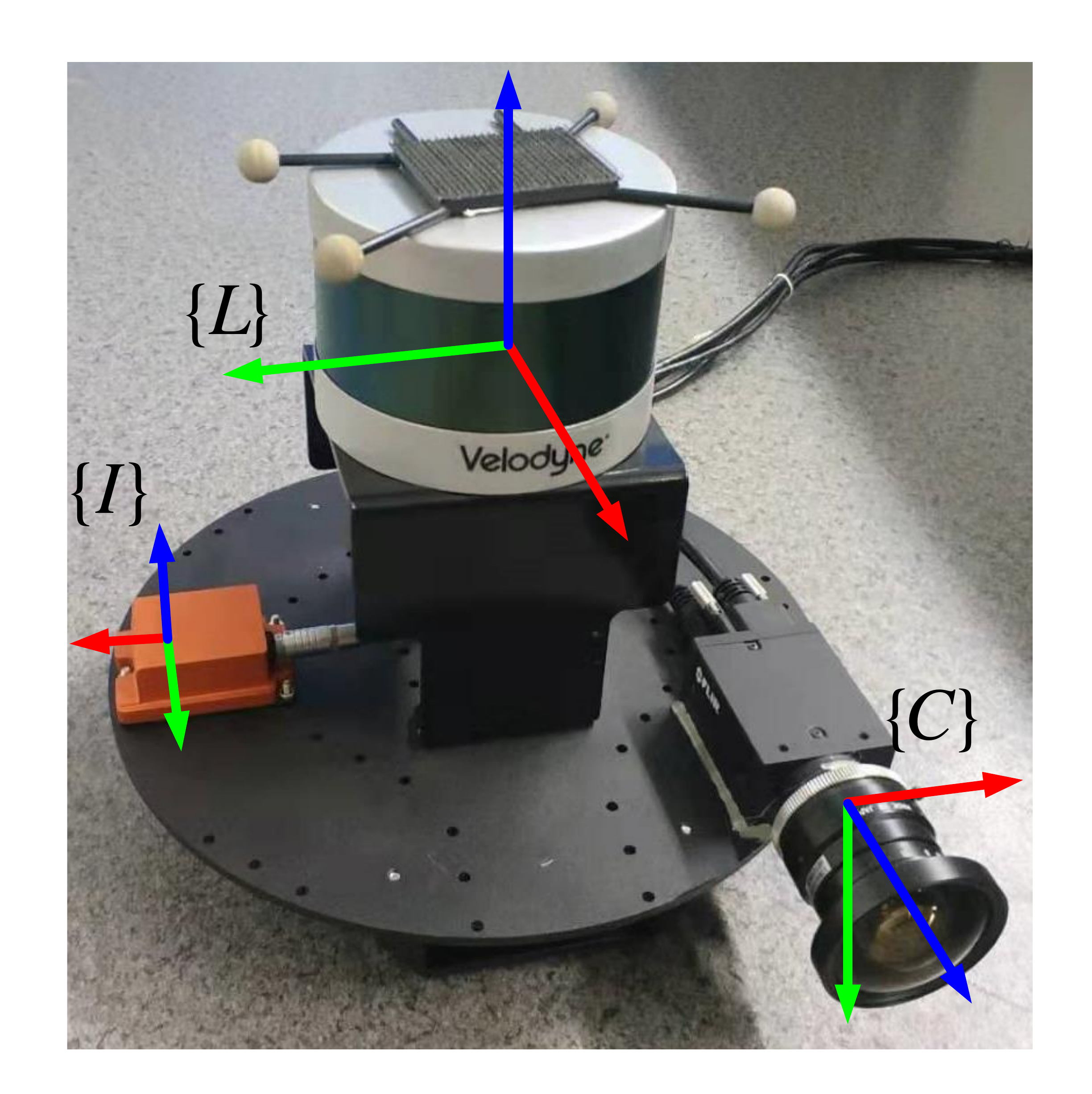}
	\end{subfigure}
	\begin{subfigure}{.4\textwidth}
		\includegraphics[trim=0 0 0 0,clip,width=\linewidth]{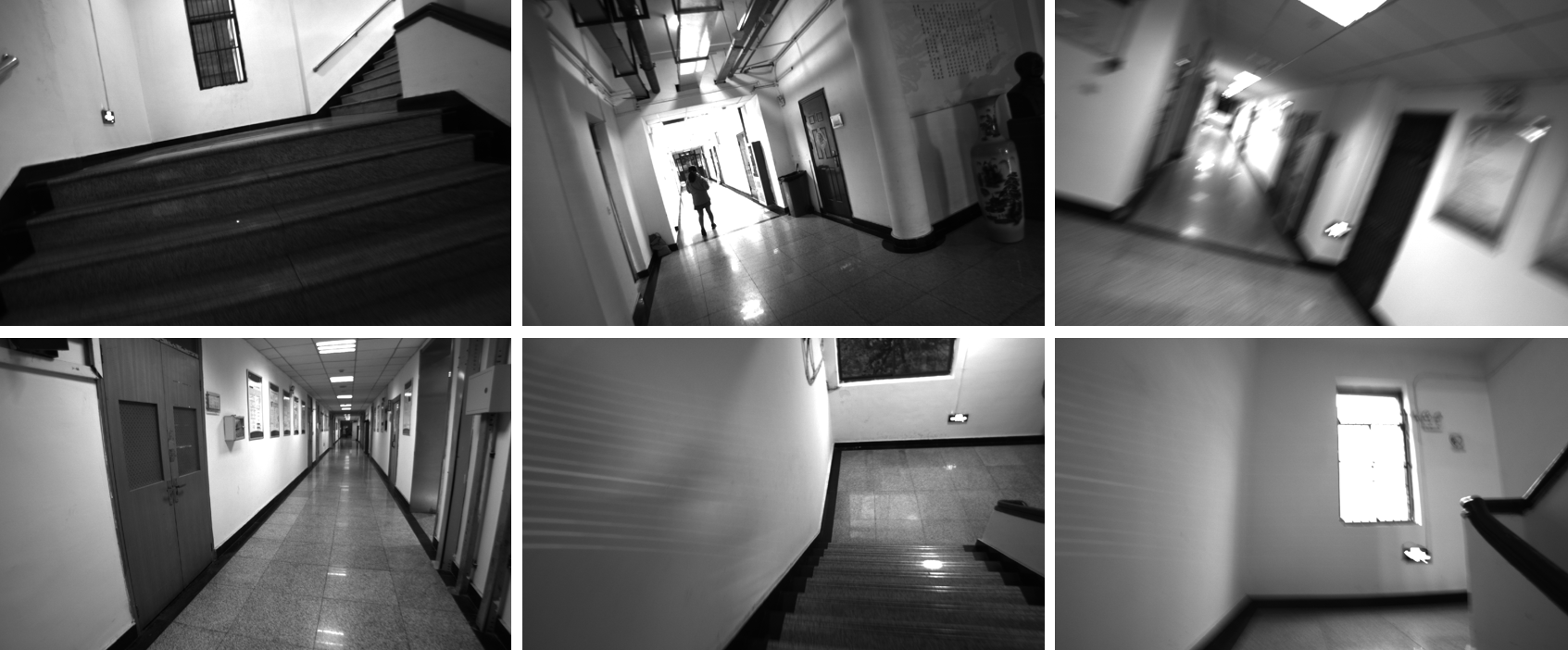}
	\end{subfigure}
	\begin{subfigure}{.26\textwidth}
		\includegraphics[trim=0 0 0 0,clip,width=\linewidth]{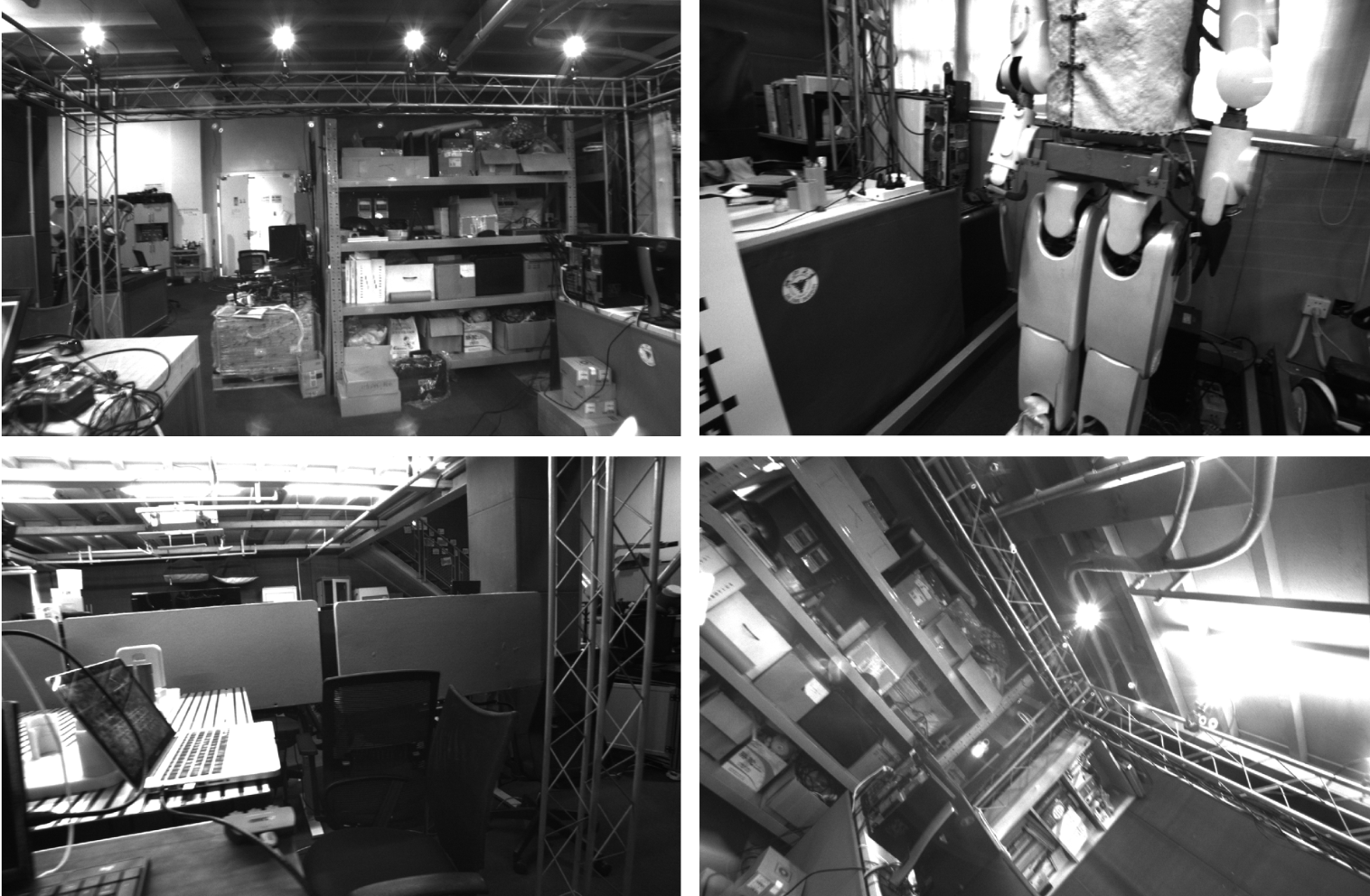}
	\end{subfigure}
	\caption{
		Left: Sensor suite with a Velodyne VPL-16, Xsens IMU, and a monocular camera. 
		Middle: Snapshots of Teaching Building sequences. 
		Right: Snapshots of Vicon Room sequences. 
	}
	\label{fig:snapshot_TB}
	\vspace*{-1.5em}
\end{figure*}

%
\begin{table}
	\centering
	\caption{
		Parameters used in our real world experiments.
	}
	\label{tab:expparams}
	\begin{adjustbox}{width=\columnwidth,center}
		\begin{tabular}{ccccc} \toprule
			\textbf{Parameter} & \textbf{Value} & \textbf{Parameter} & \textbf{Value} \\ \midrule
			Cam Freq. (hz) & 20 & IMU Freq. (hz) & 400 \\
			LiDAR Freq. (hz) & 10 & Image Res. (px) & 1920$\times$1200 \\
			Num. Clones Image & 11 & Num. Clones LiDAR & 8  \\
			Max Num. SLAM Point & 20 & Max Num. SLAM Plane  & 8 \\\bottomrule
		\end{tabular}
	\end{adjustbox}
	\vspace*{-2em}
\end{table}
We further validate the proposed LIC-Fusion 2.0. using  our multi-sensor platform that consists of a Velodyne VLP-16, an Xsens IMU, and a global-shutter monocular camera
(see Fig.~\ref{fig:snapshot_TB}).
All sensors publish asynchronously, with all time offsets estimated online with the zero as the initial guesses.
The image processing pipeline is based on our prior work OpenVINS \cite{Geneva2020ICRA}, while the LiDAR processing pipeline is proposed in this work. Note that IMU is necessary as the base sensor while, by design, the LiDAR and camera can be turned on/off without affecting performance.
Videos are recorded when generating experimental results\footnote{\url{https://www.youtube.com/watch?v=waE5nepxD-Q},\\ \url{https://drive.google.com/open?id=1cLczzQVpsgtRQhuCXAHOO563gFJSZckX}}.

\subsection{Teaching Building Sequences}
\begin{figure*}
	\centering
	\begin{subfigure}{.22\textwidth}
		\includegraphics[trim=0 0 0 0,clip,width=\linewidth]{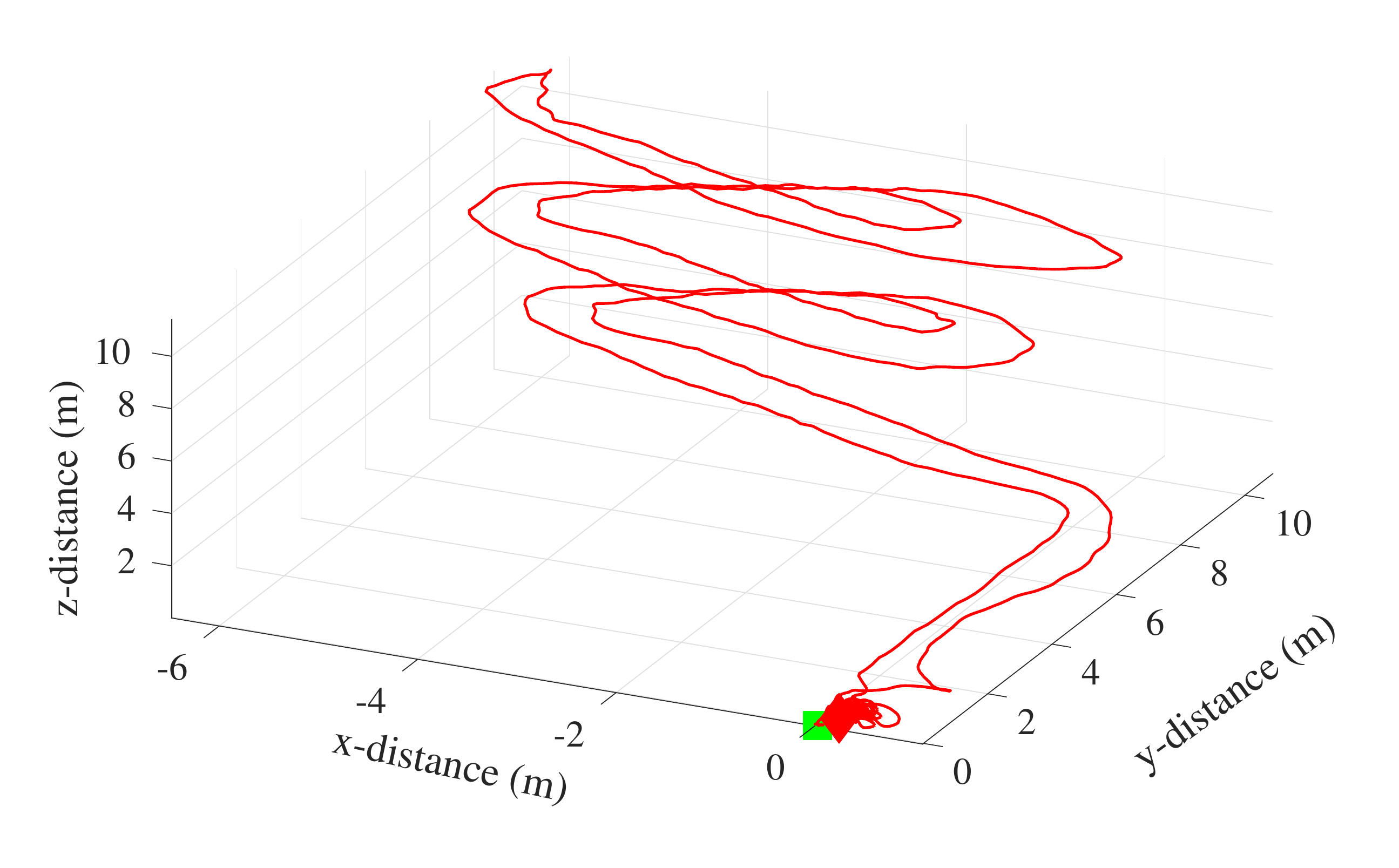}
	\end{subfigure}
	\begin{subfigure}{.22\textwidth}
		\includegraphics[trim=0 0 0 0,clip,width=\linewidth]{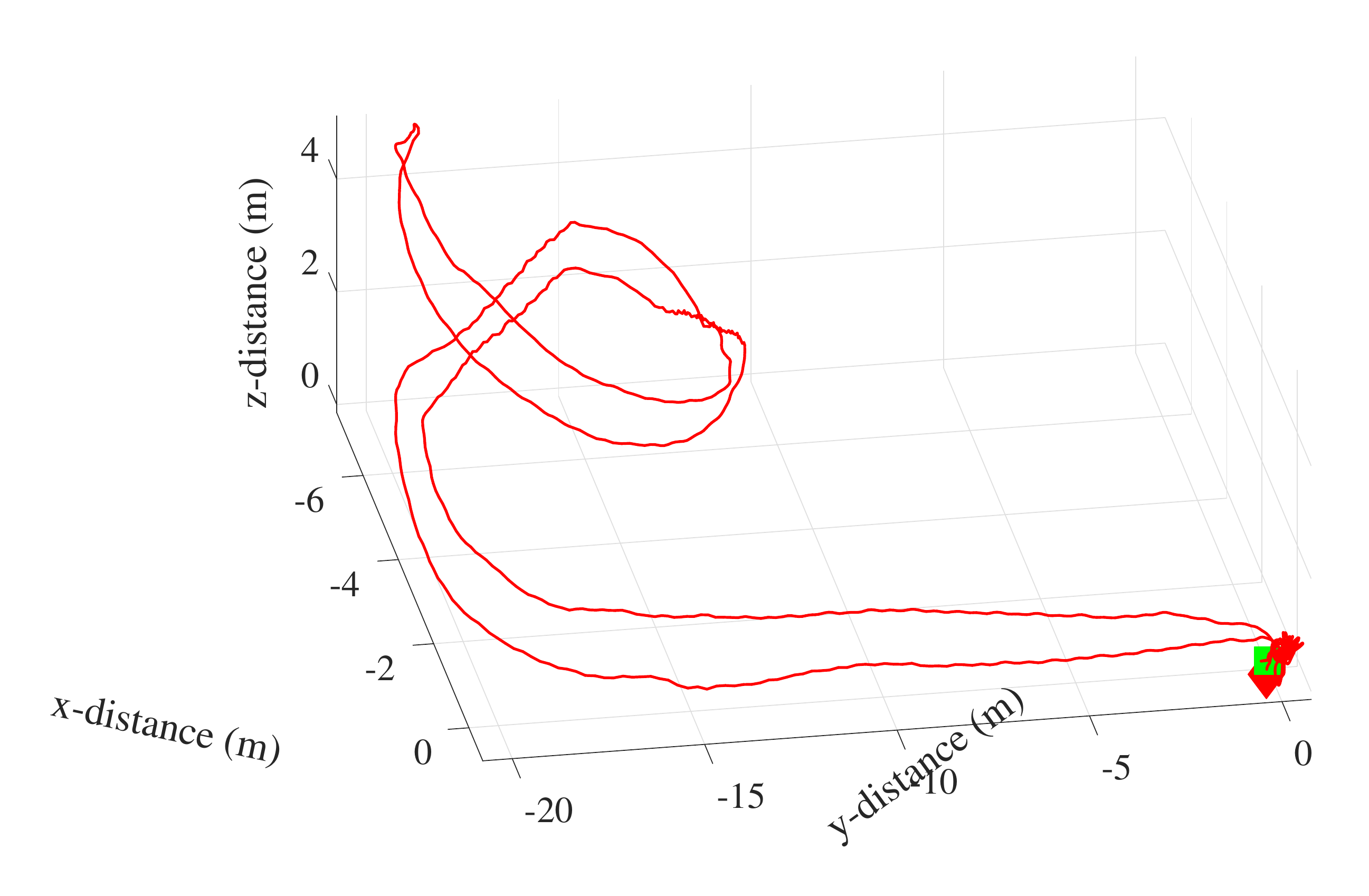}
	\end{subfigure}
	\begin{subfigure}{.17\textwidth}
		\includegraphics[trim=0 0 0 0,clip,width=\linewidth]{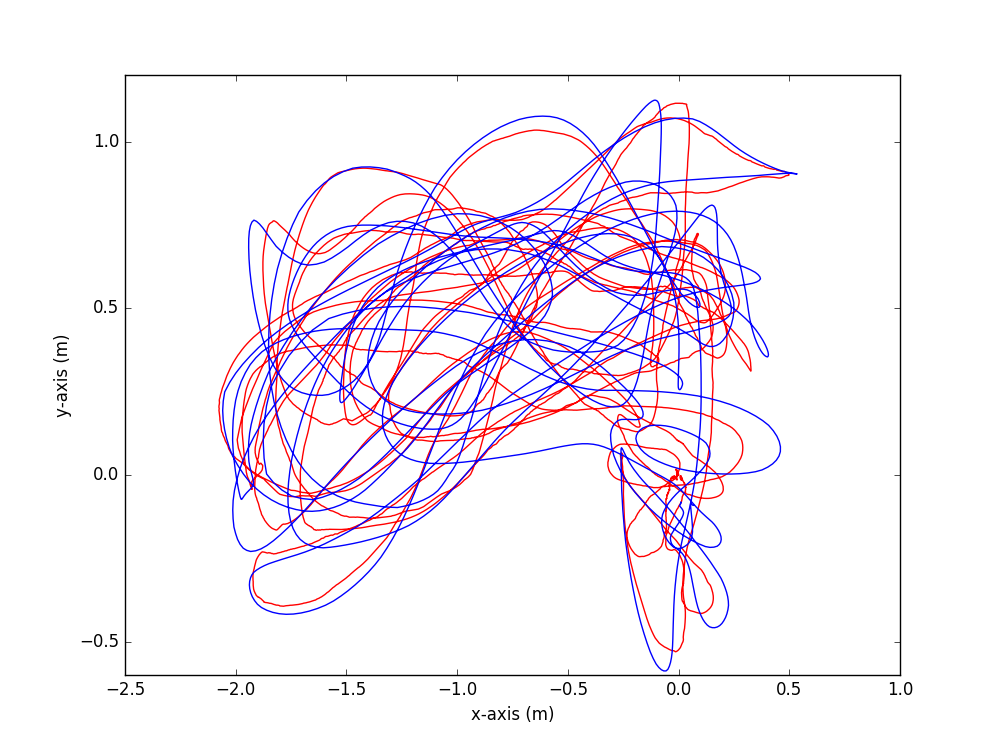}
	\end{subfigure}
	\begin{subfigure}{.32\textwidth}
		\includegraphics[trim=0 0 0 0,clip,width=\linewidth]{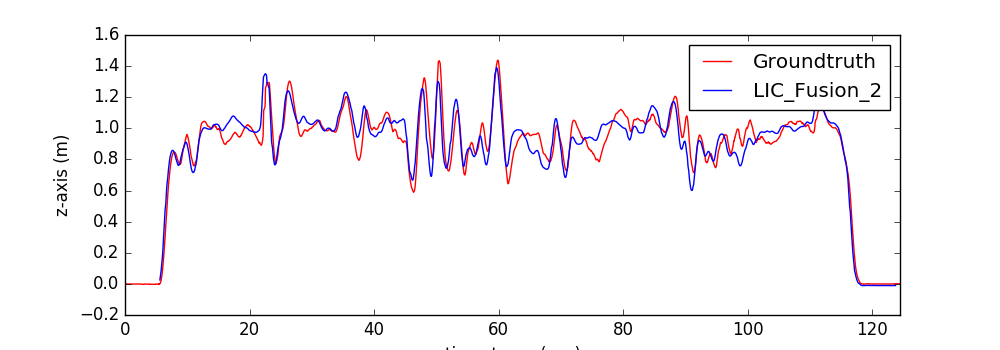}
	\end{subfigure}
	\caption{
		Left Two:
		Estimated trajectories with LIC-Fusion 2.0 on Teaching Bulding Seq 1 and 2.
		Right Two: 
		Estimated trajectories with LIC-Fusion 2.0 on Vicon Room Seq 2 overlaid with ground truth.
	}
	\label{fig:traj_tb}
\end{figure*}

%
%
The proposed system is first evaluated on data (Fig.~\ref{fig:snapshot_TB} and Table~\ref{tab:expparams}) collected within a teaching building at Zhejiang University. 
%
%
%
%
Since we started and ended in the same position when collecting data, the start-to-end drift (supposed to be zero) is used for system performance evaluation (see Fig.~\ref{fig:traj_tb}).
The averaged start and end errors of 5 runs tested on 7 sequences are shown in Table~\ref{tab:tb_exp_finaldrift}. 
In the experiments, we compare the proposed plane landmarks enhanced LiDAR-IMU-CAM odometry (\textit{LIC-Fusion 2.0}) with its subsystems (IMU-CAM system: \textit{OpenVINS}, LiDAR-IMU system: \textit{Proposed-LI}) and the other state-of-the-art algorithms, such as the LiDAR odometry
, (\textit{LOAM}~\cite{zhang2014loam}), the tightly-coupled LiDAR-Inertial odometry and mapping method (\textit{LIO-MAP}~\cite{ye2019tightly}), and our prior work (\textit{LIC-Fusion}~\cite{zuo2019lic}). 
%
%
%
Due to aggressive motion, degraded structures, lighting changes, some algorithms fail to work on certain sequences.  
In the Table~\ref{tab:tb_exp_finaldrift}, we omit severe failures marked by ``-" when the norm of final drift is larger than $30$ meters.
In Seq 1, the camera-based OpenVINS fails to track visual features due to huge camera exposure changes when we go upstairs under poor lighting conditions.
The proposed-LI subsystem has a larger drift on Seq 3 and Seq 6, in which the sensor suite traversed long corridors with only parallel planes observed.
LIO-MAP also fails on Seq 3 with long corridors even with a maintained global map. 
In general, compared to other algorithms, the proposed LIC-Fusion 2.0 is more robust and can achieve higher accuracy on most sequences. 
%
%
%
Note that in a typical indoor scenario of Seq 5, there are 18.81 MSCKF planes and 2.09 SLAM planes used for the update on average.

\begin{table*}[t]
	\renewcommand{\arraystretch}{1.2}
	\caption{
		Averaged Start-to-End drift Error of 5 runs on Teaching Building Sequences (unit meters). 
		The lengths for Seq1 - Seq 7 are around 108, 124, 237, 195, 85, 140, 83 meters, respectively.
		Note that estimated trajectories on Seq 1 and 2 are shown in Fig.~\ref{fig:traj_tb}.
	}
	\vspace*{-1.0em}
	%
	\label{tab:tb_exp_finaldrift}
	\begin{center}
		\resizebox{0.99\textwidth}{!}
		{
			\begin{tabular}{cccccccc} \toprule
				\textbf{Methods}& \textbf{Seq 1} & \textbf{Seq 2} & \textbf{Seq 3} & \textbf{Seq 4} & \textbf{Seq 5} & \textbf{Seq 6} & \textbf{seq7} \\\midrule
				LIC-Fusion 2.0 & 0.213, 0.074, 0.338 & 0.136, -0.107, -0.140  &  \textbf{0.689, -0.404, -0.172} & \textbf{0.456, 0.122, -0.322} & \textbf{0.054, -0.168, -0.027} & \textbf{0.025, -0.654, 0.199} &  1.911, 0.226, -0.166\\
				OpenVINS-IC & -,\quad  -,\quad - & -1.765,-1.149,-0.836& 3.917, 3.552, -0.475& 3.181, -0.595, -1.372 & -1.093,-0.083,-0.362 & -0.085,-3.223,-0.143 & -2.312, 1.562, 0.247\\
				Proposed-LI & 0.401, -0.195, 0.655 & 0.203, 0.503, 0.037&-,\quad  -,\quad - & 0.164,22.251,0.502 & 1.542, -2.110, 0.342 &-,\quad  -,\quad - & 1.242, -0.462, -0.530\\
				LOAM & 0.831, -5.145, -0.607& \textbf{-0.059, -0.065, 0.073} & -3.418, 3.938, -21.364& -0.933, -8.395, 0.098& -9.014, 1.084, -0.300 &
				-0.130, 0.461, 2.960 & 1.612, 0.000, -2.867 \\
				LIO-MAP& \textbf{-0.104, 0.057, 0.092}& -0.019, -0.423, 0.223&-,\quad  -,\quad -& 0.471, -0.215, -1.37&  0.147, 0.017, -0.232& 0.206, 0.125, 1.530 & \textbf{0.019, -0.039, -0.142}\\
				LIC-Fusion& -0.740, 0.0401, 0.222 & 0.293, 0.984, -0.656& 1.216, 1.831, -0.465& -1.117, 0.607, 0.529& -0.382, -2.248, -0.905& -3.295, -1.934, 0.585& -0.912, -0.847, 0.377 \\
				\bottomrule
			\end{tabular}
		}
	\end{center}
	\vspace*{-1.5em}
\end{table*}

\subsection{Vicon Room Sequences}

\begin{table*}[t]
	\renewcommand{\arraystretch}{1.0}
	\caption{
		Averaged ATE of 5 runs on Vicon Room Sequences (units degrees/meters). The lengths for Seq 1 - Seq 6 are  42.62, 84.16, 33.92, 53.14, 49.74, 87.87 meters, respectively.
		Note that estimated trajectory on Seq 2 is shown in Fig.~\ref{fig:traj_tb}
	}
	\vspace*{-1.0em}
	%
	\label{tab:ate_104data}
	\begin{center}
		\resizebox{0.80\textwidth}{!}{
			\begin{tabular}{cccccccc} \toprule
				\textbf{Methods}& \textbf{Seq 1} & \textbf{Seq 2} & \textbf{Seq 3} & \textbf{Seq 4} & \textbf{Seq 5} & \textbf{Seq 6} & \textbf{Average} \\\midrule
				LIC-Fusion 2.0  & 2.537 / 0.097 & 1.870 / 0.145 & \textbf{1.940 }/ \textbf{0.101} & \textbf{2.081} / 0.116 & \textbf{2.710 }/ 0.104 & \textbf{3.320 }/\textbf{ 0.113} & \textbf{2.410} / \textbf{0.113} \\
				OpenVINS-IC & 2.625 / \textbf{0.094} & \textbf{1.741} / 0.177 & 3.131 / 0.273 & 2.404 / \textbf{0.115} & 2.962 / 0.129 & 3.953 / 0.129 & 2.803 / 0.153 \\
				Proposed-LI & 2.333 / 0.199 & 3.325 / 0.444 & 2.810 / 0.306 & 5.335 / 0.272 & 3.332 / 0.440 & 4.866 / 0.412 & 3.667 / 0.345 \\
				LOAM & 5.880 / 0.156 & 6.414 / \textbf{0.134} & 15.384 / 0.333 & 6.354 / 0.150 & 5.542 / 0.140 & 7.095 / 0.188 & 7.778 / 0.183 \\
				LIO-MAP & - / - & 5.608 / 0.214 & - / - & - / - & 4.890 / 0.170 & 12.862 / 0.238 & 7.786 / 0.207 \\
				LIC-Fusion & \textbf{2.345} / 0.097 & 1.879 / 0.173 & 1.973 / 0.104 & - / - & 2.743 / \textbf{0.100} & 3.788 / 0.131 & 2.546 / 0.121 \\
				\bottomrule
			\end{tabular}
		}
	\end{center}
	\vspace*{-1.5em}
\end{table*}

Data sequences collected within a VICON are also used for system evaluation. Clutters in the environment (shown in Fig.~\ref{fig:snapshot_TB}) pose challenges for data associations of LiDAR points.
The averaged ATE~\cite{zhang2018tutorial} are computed with the provided ground truth to compare the LIC-Fusion 2.0, OpenVINS-IC, Proposed-LI, LOAM, LIO-MAP, and LIC-Fusion. 
%
%
The results are shown in Table.~\ref{tab:ate_104data} and Fig.~\ref{fig:traj_tb}, the cases with transitional errors more than 20 meters are marked with ``-".
%
%
The proposed LIC-Fusion 2.0 with reliable data associations over the sliding window outperforms the other algorithms.
%
We appreciate the help from the authors of LIO-MAP~\cite{ye2019tightly} for parameters tuning to achieve better accuracy. However, LIO-MAP still fails on some sequences due to error-prone data association in clutter environment and lack of time synchronization between LiDAR and IMU. 

\begin{figure*}
	\centering
	\begin{subfigure}{.32\textwidth}
		\includegraphics[trim=0 0 0 0,clip,width=\linewidth]{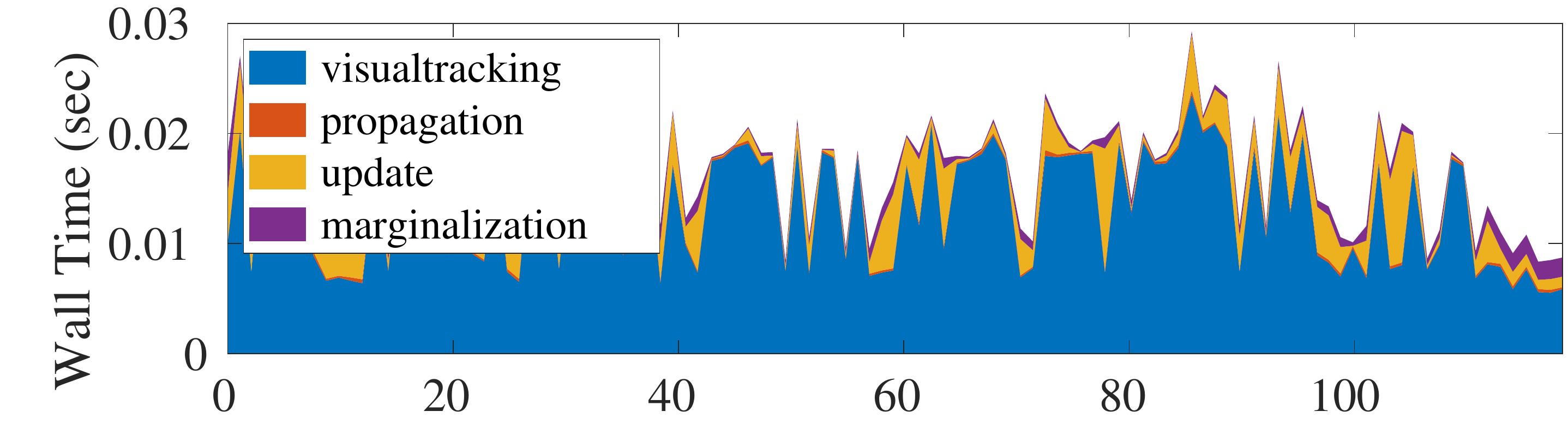}
	\end{subfigure}
	\begin{subfigure}{.32\textwidth}
		\includegraphics[trim=0 0 0 0,clip,width=\linewidth]{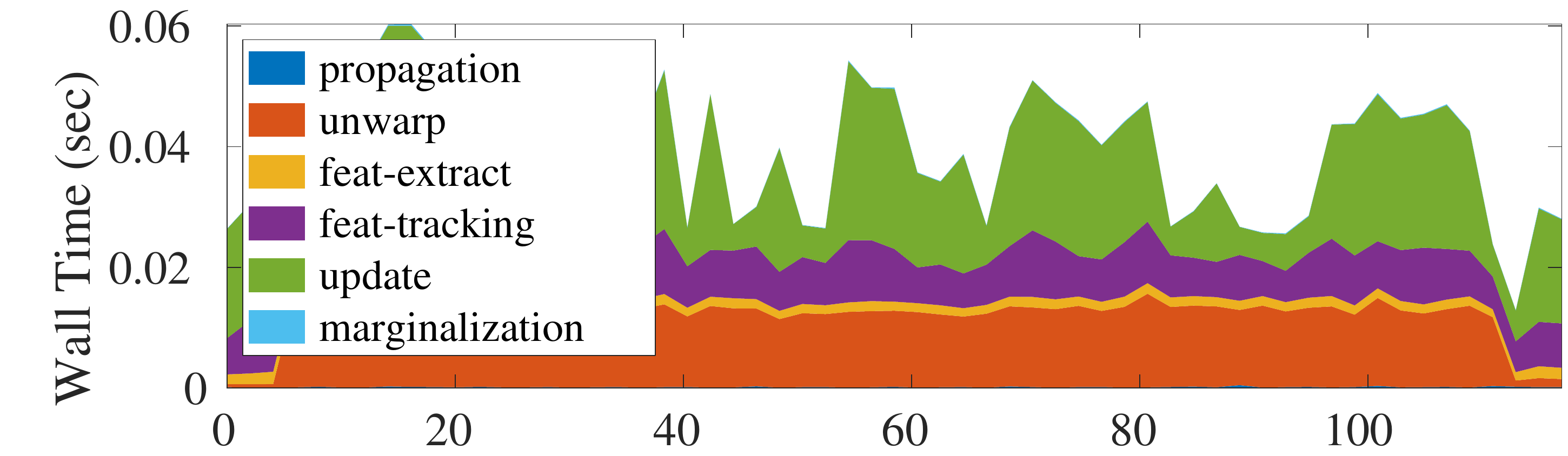}
	\end{subfigure}
	\begin{subfigure}{.32\textwidth}
		\includegraphics[trim=0 0 0 0,clip,width=\linewidth]{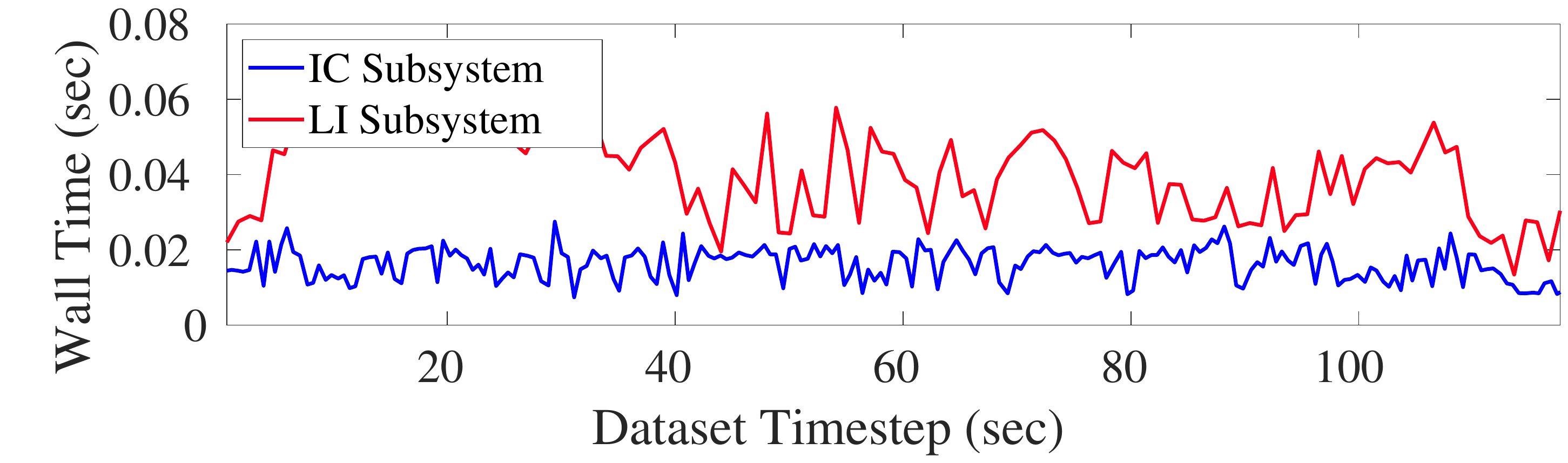}
	\end{subfigure}
	\caption{The run time analysis of the proposed system. 
	}
	\label{fig:runtime}
	\vspace*{-2.0em}
\end{figure*}

%
%
%
%
The results demonstrate that LIC-Fusion 2.0 with the novel temporal plane tracking and online spatial/temporal calibration can achieve better accuracy than existing LiDAR-IMU-CAM fusion algorithms. 
%
We further examine the computational cost (shown in Fig.~\ref{fig:runtime}) of the main stages when running it on Seq 6 on a desktop computer with Intel i7-8086k CPU@4.0GHz.
The averaged processing time for its IMU-CAM subsystem is 0.0168 seconds, and for its LiDAR-IMU subsystem is 0.0402 seconds.
Thus LIC-Fusion 2.0 is suitable for real-time applications in this indoor scenario.
%

\section{Conclusions and Future Work}

In this paper, we have developed a robust and efficient sliding-window plane-feature tracking algorithm to process 3D LiDAR point cloud measurements.
We integrated this tracking algorithm into our prior  LIC-Fusion estimator resulting in LIC-Fusion 2.0 with improved performance.
%
%
In particular, during the proposed plane-feature tracking, 
we have advocated a new outlier rejection criteria to improve feature matching quality by taking to account the uncertainty of the LiDAR frame transformations.
Additionally, we have investigated the observability properties of the linearized LIC system model in-depth and identified the degenerate cases for spatial-temporal LiDAR-IMU calibration with plane features. 
The proposed approach has been validated in both simulated and real-world datasets and shown to achieve better accuracy than the state-of-the-art algorithms.
In the future, sliding-window edge-feature tracking in LiDAR scans will be investigated.

{
\bibliographystyle{IEEEtran}  
\bibliography{root}

}

\end{document}